\documentclass[letterpaper,10pt,journal]{IEEEtran}

\usepackage{amssymb}
\usepackage{amsthm}
\usepackage{lineno}
\usepackage{amsmath,amsfonts}
\usepackage{gensymb}
\usepackage{algorithm,algpseudocode}
\usepackage{array}
\usepackage[caption=false,font=normalsize,labelfont=sf,textfont=sf]{subfig}
\usepackage{textcomp}
\usepackage{stfloats}
\usepackage{url}
\usepackage{verbatim}
\usepackage{graphicx}
\usepackage{balance}
\usepackage{afterpage}
\usepackage{tabularx}
\usepackage[colorlinks,bookmarksopen,bookmarksnumbered,citecolor=red,urlcolor=red]{hyperref}
\usepackage[noadjust]{cite}
\usepackage{booktabs}
\usepackage{makecell}

\begin{document}

\title{Adaptive LiDAR Odometry and Mapping for Autonomous Agricultural Mobile Robots in Unmanned Farms}

\author{Hanzhe Teng, Yipeng Wang, Dimitrios Chatziparaschis, Konstantinos Karydis%
\thanks{*We gratefully acknowledge the support of NSF \# IIS-1901379, \# CMMI-2046270, and \# CMMI-2326309, USDA-NIFA \# 2021-67022-33453, ONR \# N00014-18-1-2252, and The University of California under grant UC-MRPI M21PR3417. Any opinions, findings, and conclusions or recommendations expressed in this material are those of the authors and do not necessarily reflect the views of the funding agencies.}
\thanks{*Corresponding author: Konstantinos Karydis.}
\thanks{Hanzhe Teng, Yipeng Wang, Dimitrios Chatziparaschis, and Konstantinos Karydis are with Department of Electrical and Computer Engineering, University of California-Riverside, Riverside, CA, USA. {\tt\small \{hteng007, ywang1040, dchat013, karydis\}@ucr.edu}}%
}

\maketitle
\thispagestyle{empty}
\pagestyle{empty}

\begin{abstract}
Unmanned and intelligent agricultural systems are crucial for enhancing agricultural efficiency and for helping mitigate the effect of labor shortage. 
However, unlike urban environments, agricultural fields impose distinct and unique challenges on autonomous robotic systems, such as the unstructured and dynamic nature of the environment, the rough and uneven terrain, and the resulting non-smooth robot motion. 
To address these challenges, this work introduces an adaptive LiDAR odometry and mapping framework tailored for autonomous agricultural mobile robots operating in complex agricultural environments. 
The proposed framework consists of a robust LiDAR odometry algorithm based on dense Generalized-ICP scan matching, and an adaptive mapping module that considers motion stability and point cloud consistency for selective map updates. 
The key design principle of this framework is to prioritize the incremental consistency of the map by rejecting motion-distorted points and sparse dynamic objects, which in turn leads to high accuracy in odometry estimated from scan matching against the map. 
The effectiveness of the proposed method is validated via extensive evaluation against state-of-the-art methods on field datasets collected in real-world agricultural environments featuring various planting types, terrain types, and robot motion profiles. 
Results demonstrate that our method can achieve accurate odometry estimation and mapping results consistently and robustly across diverse agricultural settings, whereas other methods are sensitive to abrupt robot motion and accumulated drift in unstructured environments. 
Further, the computational efficiency of our method is competitive compared with other methods. 
The source code of the developed method and the associated field dataset are publicly available at \url{https://github.com/UCR-Robotics/AG-LOAM}.
\end{abstract}



\section{Introduction}
A critical and fundamental component in unmanned farms is the ability to perform accurate localization and mapping in agricultural field settings. 
On one hand, accurate localization serves as the basis for autonomous navigation of agricultural robots~\cite{malavazi2018lidar-only, blok2019navigation, hiremath2014laser_nav, opiyo2021medial} and supports fully automated in-field operations, such as pesticide spraying~\cite{meshram2022pesticide}, leaf sample retrieval~\cite{merrick2022iros}, and collection of various other sources of information~\cite{merrick2021case, Amel2023ram, gene2020fruit, ji2021obstacle}.
On the other hand, acquisition of detailed and accurate maps of crops on a regular basis can provide growers and agronomists with information about plant growth~\cite{zhang2012crop_height, dhami2020crop_height, kim2021stereo_crop_height,chatz2024go}) and enable monitoring of dynamic changes in the field~\cite{bietresato2016crop_monitoring, weiss2011plant,chatz2024robot}.
When multispectral cameras are employed, such maps can further provide information on plant water stress level and crop health conditions~\cite{underwood2016mapping, pretto2020building,chatz2024go}.

To develop such techniques, Light Detection and Ranging (LiDAR) technology has been increasingly applied in the agriculture domain, owing to recent advances in enhancing computation at edge devices and decreasing hardware costs~\cite{lin2015lidar-review, rivera2023lidar-review}. 
Ubiquitous among past efforts on robot navigation in agricultural environments was the reliance on external infrastructure such as GNSS and fiducial markers, which might not be accessible at all times~\cite{rovira2015gnss}. 
A recent alternative approach is to navigate by leveraging only (or mostly) onboard sensors, including LiDAR, cameras, Inertial Measurement Units (IMUs), etc. 
Among these sensors, LiDAR offers superior accuracy and reliability~\cite{rivera2023lidar-review} and is capable of providing a detailed 3D reconstruction of the environment~\cite{dong2017recon, marks2022recon}. 
Thus, this work aims to develop a LiDAR-based localization and mapping framework to address the following two problems simultaneously: 1) providing real-time odometry to support the autonomous navigation of agricultural robots, and 2) obtaining accurate maps of agricultural fields to enable regular crop monitoring.

\begin{figure}[!t]
\centering
\includegraphics[width=\linewidth]{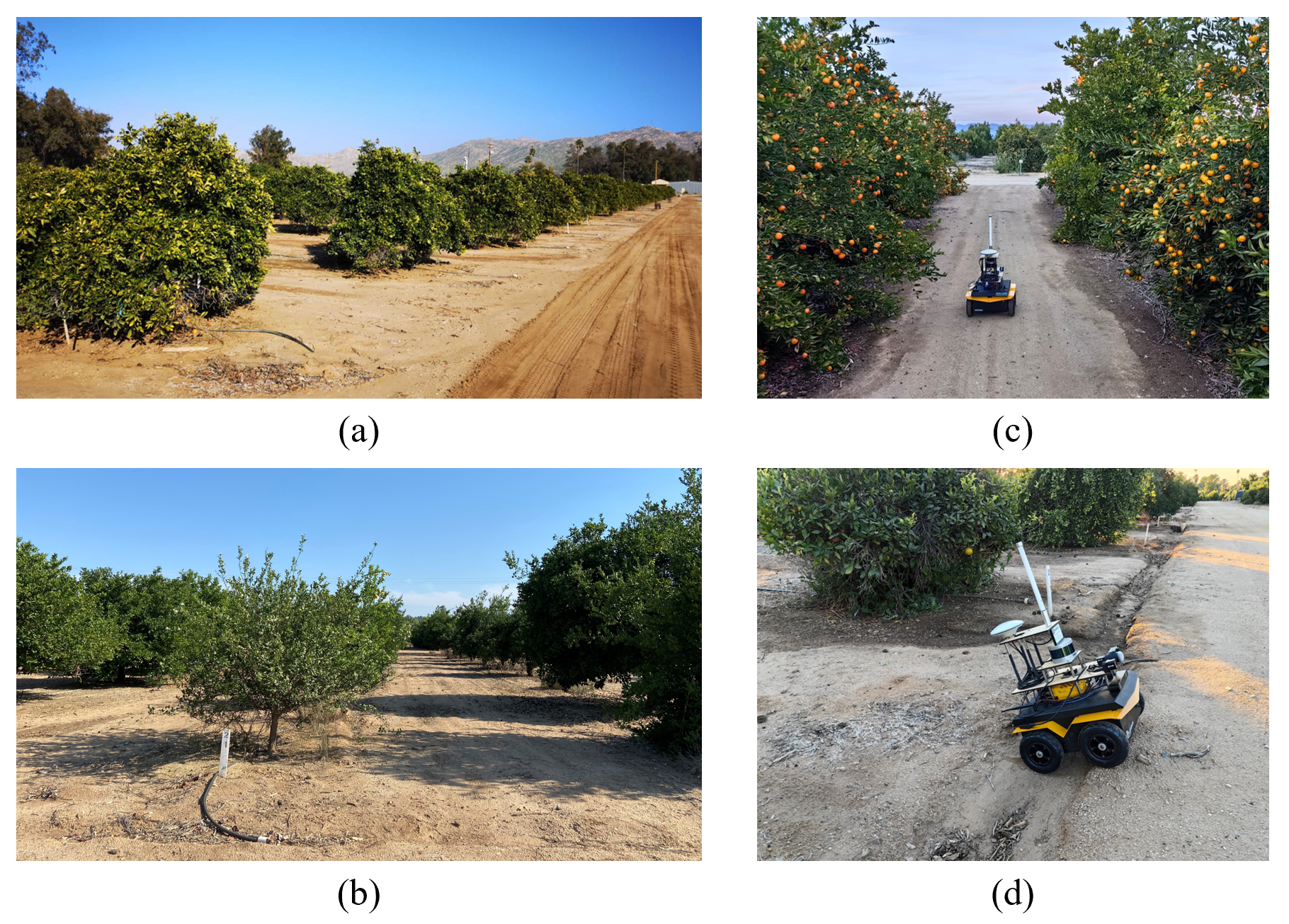}
\caption{Assorted views from the agricultural fields at the University of California, Riverside, where experiments were conducted. (a) A sample field featuring flat ground and navel orange trees planted in uniform, squared-shape pattern. (b) A sample field featuring a furrow on the side and a mixture of citrus trees planted in-row, with some space in between. (c) The Clearpath Robotics Jackal mobile robot used in this work shown while navigating through a field of densely in-row planted citrus trees. (d) The Jackal mobile robot captured at an instance of traversing a furrow while navigating in the field.}
\label{fig_field}
\end{figure}

However, unlike urban environments that LiDAR-based algorithms have been studied fairly extensively (e.g., ~\cite{behley2018suma, park2018elastic, vizzo2021poisson, dellenbach2022cticp}), agricultural fields impose distinct and unique challenges on autonomous robotic systems.
Some key challenges include the unstructured operating environment, the dynamic nature of some parts of the environment (e.g., trees swaying due to wind), as well as the rough and uneven terrain (such as rocks, bumps and dips) resulting in less smooth robot motion.
These are essentially distinct sources of uncertainty that can lead to inconsistent, unreliable and motion-distorted point cloud data acquired by the LiDAR sensor.

To address the aforementioned challenges, existing methods leverage a variety of sensing modalities and fuse them with LiDAR data for joint estimation. 
The predominant technique involves incorporating an IMU for ego-motion estimation and distortion correction, spanning both tightly-coupled~\cite{ye2019lio-mapping, xu2021fastlio, xu2022fastlio2} and loosely-coupled approaches~\cite{palieri2020locus}.
However, requiring specific combinations of sensors can restrict the application of proposed algorithms to broader scenarios because of limited hardware availability or an increased burden on calibrating sensors both spatially and temporally. 
Additionally, the inclusion of multiple sensors often requires substantial system integration efforts~\cite{teng2023multimodal}, which in turn require specific technical expertise to both develop and maintain. 
This may reduce the chances for sensor-based agricultural robotics technology to be adopted in practice~\cite{gil2023low}. 

An alternative approach, which is also the focus of this work, is to improve odometry and localization algorithms to achieve comparable performance using \emph{a single LiDAR sensor}. 
This is a challenging problem because of the potential unawareness of motion distortion and/or dynamic objects.
Although there exist several methods designed for LiDAR-only applications in indoor or outdoor structured environments (e.g.,~\cite{behley2018suma, shan2018lego, pan2021mulls, wang2021floam, dellenbach2022cticp,vizzo2023kiss-icp}), few methods focus on outdoor unstructured environments such as agricultural fields.
A sample agricultural environment considered in this work is shown in Figure~\ref{fig_field}.

To this end, this work focuses on the development of a new framework to address the unique set of challenges in odometry and mapping for autonomous robots operating in agricultural environments.
The proposed framework consists of a robust LiDAR odometry algorithm that performs dense scan matching on incoming consecutive point clouds and an adaptive mapping module that strategically updates the map according to robot motion stability and point cloud consistency.
The core design principle of this framework is to retain only incrementally consistent points on the map and reject as many unreliable points as possible.
This can in turn lead to high reliability in robot position and orientation (pose) estimated from scan matching against the map. 
The effectiveness of the proposed method is validated via extensive evaluation against state-of-the-art methods on different datasets collected from real-world agricultural fields featuring various planting environments and robot motion profiles. 
The source code of our method and the datasets collected in the field are publicly available at \url{https://github.com/UCR-Robotics/AG-LOAM}.

\section{Related Works}
We briefly review LiDAR odometry and mapping algorithms related to our proposed method. In summary, such algorithms can be grouped into two main categories: 1) feature-based methods, and 2) dense methods.

\textbf{Feature-based Methods.}
Feature-based methods aim to extract various types of local features and establish correspondences via feature matching.
LOAM~\cite{zhang2014loam} proposed to classify features into sharp points and flat surfaces, and then optimize point-to-line and point-to-plane distances between matched features. 
Subsequent works have built upon the LOAM framework to reduce computational cost~\cite{wang2021floam}, incorporate segmentation and ground features~\cite{shan2018lego}, integrate semantic segmentation~\cite{chen2020sloam}, and make it adaptive to small field of view LiDARs~\cite{lin2020loamlivox}.
Generally, feature-based methods are fast and lightweight but may be sensitive to the consistency of feature extraction and the accuracy of feature matching.

\textbf{Dense Methods.}
In contrast to feature-based methods, dense methods directly make use of all points without a clear distinction of feature types, such as the Iterative Closest Point (ICP) method~\cite{besl1992icp} or the Normal Distribution Transform (NDT) method~\cite{magnusson2009ndt3d}. 
Correspondences are established by nearest neighbor search in 3D space and residual errors are computed for each established pair of points in each iteration. 
Various works have built on top of the ICP method, including point-to-point ICP~\cite{vizzo2023kiss-icp}, multi-metric ICP~\cite{pan2021mulls}, elastic ICP~\cite{park2018elastic, dellenbach2022cticp} and scan-to-model ICP~\cite{behley2018suma, deschaud2018imls-slam, chen2019suma++}.

Among these works, one notable variant is the Generalized ICP (GICP) method proposed by~\cite{segal2009gicp}, where a probabilistic model is incorporated into the computation of error metrics.
By applying various regularization techniques to the eigenvalues of covariance matrices, GICP can act as point-to-point, point-to-plane, or plane-to-plane ICP.
Two notable works in this vein are HDL-Graph-SLAM~\cite{koide2019hdl} and LOCUS~\cite{palieri2020locus, reinke2022locus}. 
HDL-Graph-SLAM is based on Fast-GICP~\cite{koide2021fast_gicp} scan matching, with the capability to integrate several types of constraints in pose graph optimization (e.g., loop closure, floor plane, IMU events), whereas LOCUS is a LiDAR-centric multi-sensor framework that relies on IMU for point cloud undistortion and takes estimates from other sensing sources as GICP seeds (initial alignments). 
Both works rely on other sensing modalities to a certain degree to reach their best performance, which may limit applicability across environments when suitable sensors are not available. 
This present work is inspired by the aforementioned two methods of GICP scan matching. 
Yet, it is distinct in that our proposed method relies solely on a single 3D LiDAR sensor but can still achieve centimeter-level localization accuracy in unstructured agricultural environments. 

\section{Methods}
In this section, we introduce our proposed adaptive LiDAR odometry and mapping framework for agricultural environments, named AG-LOAM, and we present the experimental setup for evaluating our algorithm. 
The proposed algorithm (Section~\ref{sec_overview}) integrates two modules sequentially (odometry in Section~\ref{sec_odometry} and mapping in Section~\ref{sec_mapping}). 
Algorithm evaluation took place over three separate experimental phases (Section~\ref{sec_experiment}).

\begin{figure}[!t]
\centering
\includegraphics[width=\linewidth]{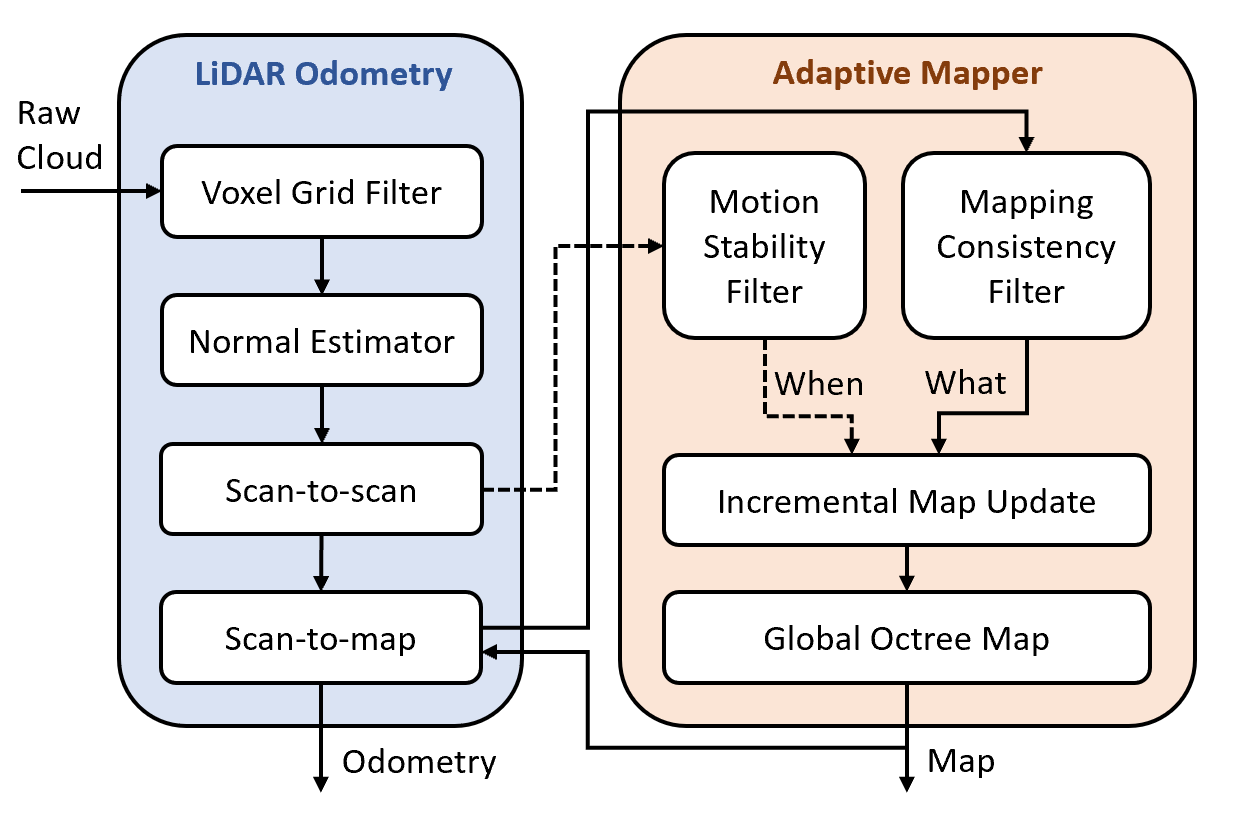}
\caption{Overall system diagram of the adaptive LiDAR odometry and mapping framework developed in this work. Solid arrows indicate the flow of point cloud data, and dashed arrows indicate information exchange between modules.}
\label{fig_framework}
\end{figure}

\subsection{Overview of the Proposed Algorithm}\label{sec_overview}
Our proposed framework (Figure~\ref{fig_framework}) consists of two primary algorithmic components: a LiDAR Odometry, and an Adaptive Mapper. 
Distinct from typical LiDAR odometry and mapping frameworks that operate concurrently in two threads (e.g.,~\cite{zhang2014loam}), the two components in our framework function in a cascaded pipeline. 
The odometry algorithm processes each incoming point cloud frame in real-time, whereas the adaptive mapper selectively updates the map only when necessary and by using fewer but more reliable points, thereby leading to minimal time overhead on the odometry process. 
Furthermore, this cascaded pipeline affords acceleration through multi-threading for tasks such as covariance estimation and nearest neighbor search queries.

To accommodate for direct integration into existing agricultural vehicles (e.g., by mounting a 3D LiDAR onto a tractor), our system does not depend on IMU data. 
Yet, if desired, an IMU data stream can be integrated in a loosely coupled way (e.g.,~\cite{palieri2020locus}) to correct motion distortion of point clouds before being processed by our proposed pipeline. 
Moreover, in line with most methods based on dense matching, our framework is optimized for 360-degree LiDAR systems that are capable of comprehensively capturing the entire surrounding environment. 
This contrasts with some solid-state LiDAR sensors, which are limited by a narrower field of view.

\subsection{LiDAR Odometry Algorithm}\label{sec_odometry}

\subsubsection{Preprocessing}
For each incoming raw point cloud frame, we first down-sample the cloud by applying a voxel grid filter.
This is a common practice in dense methods. The voxel size can be set to a fixed number (e.g., 0.1-0.5\;m, as per~\cite{koide2021fast_gicp, palieri2020locus}), or adjusted adaptively in every frame to meet a designated number of points after down-sampling so to maintain a desired computational load~\cite{reinke2022locus}. 
In our method we combine the above two approaches in a way that a fixed voxel size is maintained unless the total number of points deviates from the designated number by a certain range; in such cases, the voxel size will be adjusted accordingly. 
This strategy is implemented to maintain a similar density level for consecutive clouds while keeping computational costs low.

It is worth noting that the voxel grid filter also plays an important role in regularizing the point distribution (i.e. making the processed point cloud more uniformly distributed). 
This is critical for the covariance estimation phase of scan matching, which relies on local neighbor points for accurate covariance matrix calculations. 
Directly utilizing the original point clouds for scan matching may lead to sub-optimal results.

What follows in preprocessing is an optional step of normal estimation, which estimates a normal direction for each point using the closest 20 points in its vicinity.
This normal direction can be used to quickly reconstruct a regularized covariance matrix~\cite{reinke2022locus}, to save computational time in two-stage scan matching, where covariance matrices may be computed four times consecutively.
However, the trade-off we have observed in practice is that the point normal estimated during the current phase may not always faithfully represent the point's neighborhood once integrated into the map. 
In such cases, recalculating the normals or covariance matrices using local neighbor points from the map becomes necessary and is more effective.

\subsubsection{GICP Scan Matching}
We apply the GICP algorithm for scan matching between two point clouds. 
Let $\mathbf{T}\in SE(3)$ denote the transformation to be estimated between two point cloud frames. 
Suppose that $\mathbf{p}_i$ is a point on the target cloud $\mathcal{P}$ and $\mathbf{q}_i$ is a point on the source cloud $\mathcal{Q}$. 
We define the residual (transformation error) as $\mathbf{d}_i = \mathbf{p}_i - \mathbf{T}\mathbf{q}_i$ for each correspondence established in source and target clouds, and assume that the covariance matrix $\mathbf{C}_i$ of each point can be estimated approximately from the local neighborhood. 
In each iteration, the GICP algorithm solves for optimal $\mathbf{T}$ as
\begin{equation}
\mathbf{T} =\underset{\mathbf{T}}{\arg \min } \sum_i \mathbf{d}_i^{\top} \left(\mathbf{C}_i^\mathcal{P}+\mathbf{T} \mathbf{C}_i^\mathcal{Q} \mathbf{T}^{\top}\right)^{-1} \mathbf{d}_i \enspace.
\end{equation}

As discussed in~\cite{segal2009gicp}, if point clouds are considered locally planar, we can regularize the covariance matrices by replacing their eigenvalues with $(\epsilon, 1, 1)$, where $\epsilon$ is a small value (e.g., $0.001$). 
In this case, GICP acts as a plane-to-plane ICP. 
We use this setting for our algorithm to operate in agricultural environments.

\subsubsection{LiDAR Odometry via Two-stage Scan Matching}
We now present our complete LiDAR odometry algorithm based on two-stage GICP scan matching, applied to point clouds following the preprocessing stage.
We assume that each input point cloud is assigned by a coordinate frame $\{ k \}$, where $k$ is an incremental integer.
We take $\{ 0 \}$ as the origin of the map frame (or world frame) that coincides with the first point cloud and denote by $\mathbf{T}_{k-1, k}$ the rigid body transformation of frame $\{ k \}$ with respect to frame $\{ k-1 \}$.
Hence, LiDAR odometry solves for the best estimate of $\mathbf{T}_{0, k}$ that describes the latest robot pose with respect to the origin of the map.

We initialize $\mathbf{T}$ as an identity matrix and begin executing the LiDAR odometry algorithm (Algorithm~\ref{algorithm1}) for each incoming point cloud associated with frame $\{ k \}$. 
In the stage of scan-to-scan matching (line 3), we initialize GICP with an identity matrix and optimize for an incremental transformation $\mathbf{\Tilde{T}}_{k-1, k}$. 
This can yield a rough estimate of the robot motion, which will be passed on to serve as an initial guess in scan-to-map matching (line 4) and facilitate mapping decisions in Adaptive Mapper (to be discussed in Section~\ref{sec_mapping}). 
In the stage of scan-to-map matching (lines 5-7), we initialize GICP with the rough estimate $\mathbf{\Tilde{T}}_{k-1, k}$ provided by scan-to-scan matching, and optimize for a refined incremental transformation $\mathbf{\hat{T}}_{k-1, k}$ by matching the current frame with the closest subset of the map. 
Finally, we compose the incremental transformations computed in two stages (line 8) and integrate them into the latest odometry estimate (line 9).

\begin{algorithm}[!h]
\caption{LiDAR Odometry via Two-stage Scan Matching}\label{algorithm1}
\begin{algorithmic}[1]
\State \textbf{Input:} current cloud $\mathcal{P}_k$, previous cloud $\mathcal{P}_{k-1}$, previous odometry estimate $\mathbf{T}_{0, k-1}$
\State \textbf{Output:} current odometry estimate $\mathbf{T}_{0, k}$
\State $\mathbf{\Tilde{T}}_{k-1, k} \gets \text{GICP}(\mathcal{P}_{k}, \mathcal{P}_{k-1})$  // scan-to-scan
\State transform $\mathcal{P}_k$ by $\mathbf{\Tilde{T}}_{k-1,k}$ to obtain $\mathcal{\Tilde{P}}_{k}$
\State $\mathbf{\Tilde{T}}_{0,k} \gets \mathbf{T}_{0,k-1} \mathbf{\Tilde{T}}_{k-1, k}$
\State transform $\mathcal{P}_k$ to map frame by $\mathbf{\Tilde{T}}_{0,k}$ and find closest in-map neighbor points to form a cloud $\mathcal{P}_{nbr}$
\State $\mathbf{\hat{T}}_{k-1, k} \gets \text{GICP}(\mathcal{\Tilde{P}}_{k}, \mathcal{P}_{nbr})$  // scan-to-map
\State $\mathbf{T}_{k-1,k} \gets \mathbf{\Tilde{T}}_{k-1, k} \mathbf{\hat{T}}_{k-1, k}$
\State $\mathbf{T}_{0,k} \gets \mathbf{T}_{0,k-1} \mathbf{T}_{k-1, k}$
\end{algorithmic}
\end{algorithm}

\subsection{Adaptive Mapping Algorithm}\label{sec_mapping}
At the end of the LiDAR odometry pipeline, we need to decide 1) if it is an appropriate time to update the map, and if so, 2) what to update to the map. 
This subsection discusses the proposed method to help construct an accurate map in real time under motion distortion and sparse dynamic objects. 

\subsubsection{Map Management}
We maintain an Octree data structure for the global point cloud map, which can support fast search and insertion operations in logarithmic time complexity (with respect to map size). 
Our implementation is based on the Point Cloud Library~\cite{rusu2011pcl}, and the nearest neighbor search operation is further sped up by multi-threading.\footnote{~We used OpenMP (\url{https://www.openmp.org/}) in this work, but other multi-threading implementations are applicable.}

We employ an adaptive policy for updating the map, based on two criteria: 1) the robot has moved (translated) a reasonably long distance (e.g., 1 m) relative to the last map update position, and 2) the current point cloud frame and the previous frame both pass the Motion Stability Filter (discussed next). 
This design can reduce unnecessary map updates and prevent the candidate frame from large rotation-induced distortion.
Once selected as the key frame for map update, the current point cloud frame will be sent to the Mapping Consistency Filter to further remove unreliable points before being inserted into the map.
Overall, the main insight is to ensure every map update is as far from distortion and unreliable points as possible before it is inserted into the map, which can in turn lead to high accuracy in odometry estimated from scan matching against the map.
We discuss these two modules in detail in the following subsections and report their ablation study results in Section~\ref{sec_ablation}.

\subsubsection{Motion Stability Filter}
\begin{figure}[h]
\centering
\includegraphics[width=0.9\linewidth]{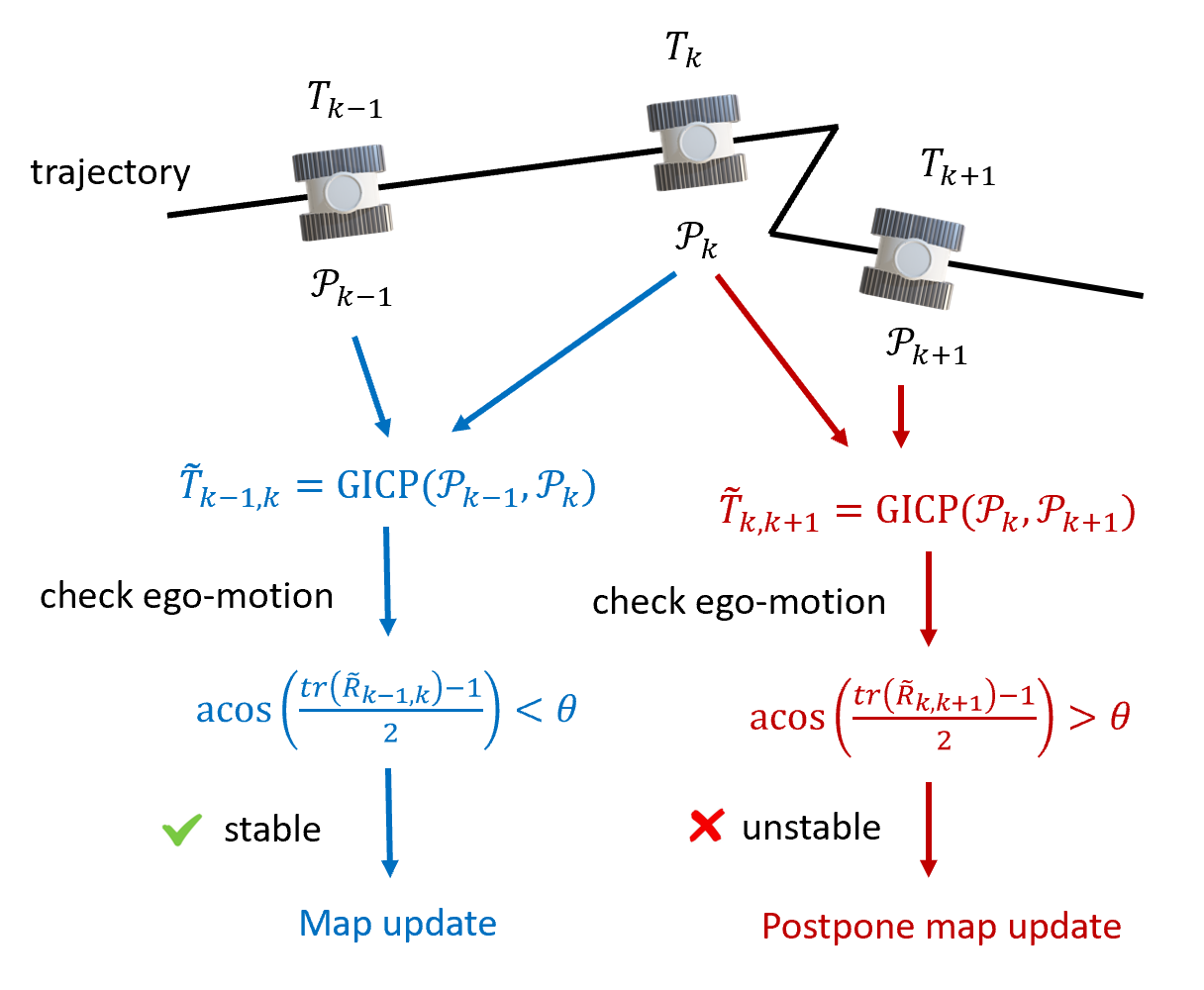}
\caption{
An illustration of the operation principle of the Motion Stability Filter.
In this example, map updates are permitted when the robot moves smoothly from $\mathbf{T}_{k-1}$ to $\mathbf{T}_{k}$. However, updates are rejected when the robot moves from $\mathbf{T}_{k}$ to $\mathbf{T}_{k+1}$, where an abrupt motion change is detected.
}
\label{fig_stability}
\end{figure}

The Motion Stability Filter determines when to update the map. The goal is to circumvent the effect of rotation-induced distortion and find an appropriate time when the robot operates with stable ego-motion. 
As illustrated in Figure~\ref{fig_stability}, a point cloud frame is considered to pass the Motion Stability Filter (or considered \textit{stable}) if the scan-to-scan matching of the current frame with respect to the previous frame presents a small-degree (e.g., 2 deg) rotation estimate. 
We denote $\theta$ as the maximum permissible rotation angle for map updates.
The key rationale behind this policy is that as the mobile robot turns in place (or follows a tight circular path), the majority of the environment remains observable via the 360-degree LiDAR sensor. 
Therefore, immediate map updates are not deemed necessary and can be postponed until the robot's motion is stable.

Note that the above criterion is a straightforward implementation specifically designed to detect large rotations. However, a variety of sanity checks can be performed on the estimated incremental ego-motion $\mathbf{T}_{k-1, k}$, including but not limited to thresholding translation vectors to detect abrupt displacements or applying swing-twist decomposition to detect rotation along a designated axis.
The key message we want to convey herein is that map updates do not need to be performed at a fixed frequency. Instead, they can be governed by an adaptive policy that takes into account the specific robot type, its dynamics, and the environment (e.g., terrain conditions). 
This adaptive approach allows for more efficient and context-aware map updates, optimizing performance under varying conditions.

\begin{figure}[h]
\centering
\includegraphics[width=0.9\linewidth]{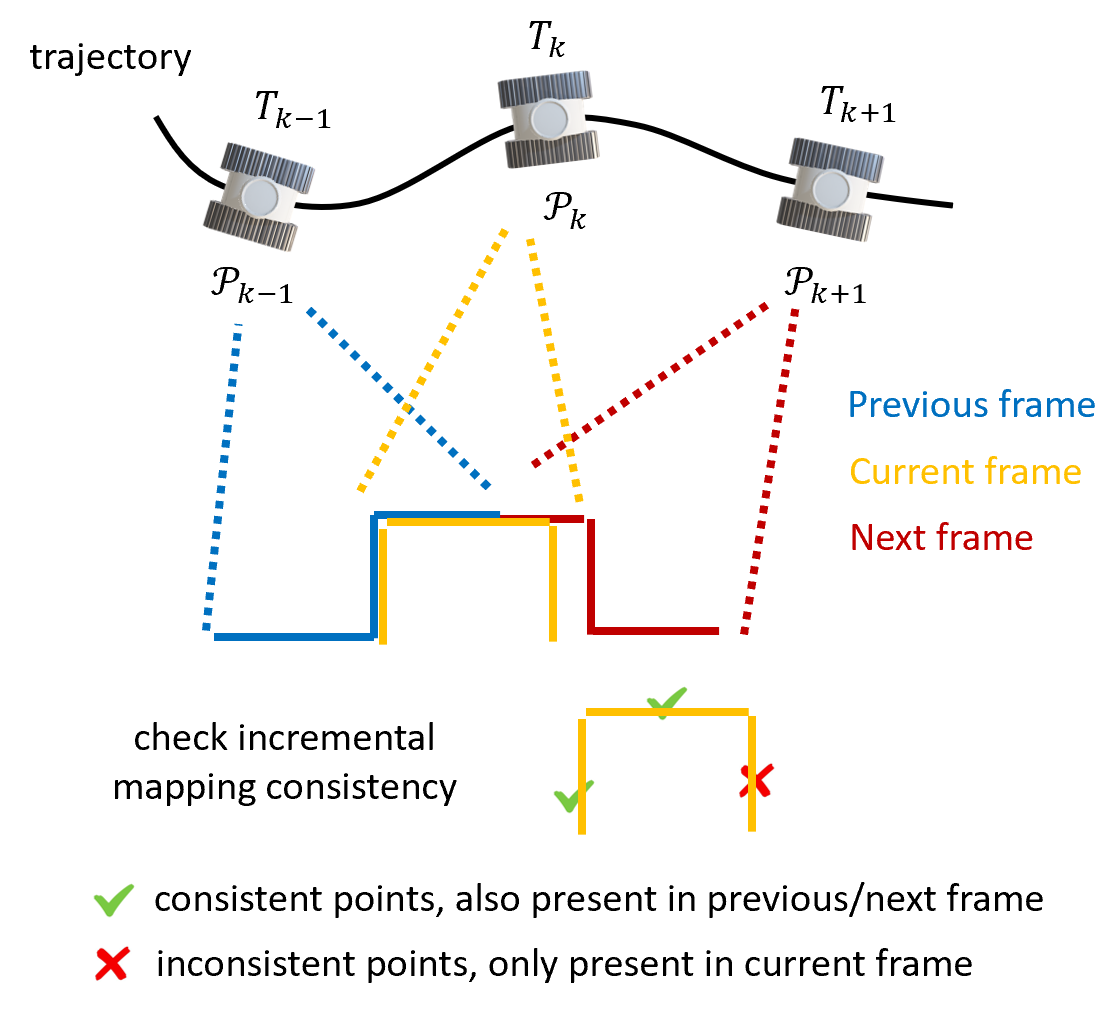}
\caption{
An illustration of the operation principle of the Mapping Consistency Filter.
In this example, points on the left and top line segments are selected because they are in close proximity to either the previous or the next frame. However, points on the right line segment are considered inconsistent and are rejected, as they only appear once in the current frame.
}
\label{fig_consistency}
\end{figure}

\begin{algorithm}[h!]
\caption{Mapping Consistency Filter}\label{algorithm2}
\begin{algorithmic}[1]
\State \textbf{Input:} current cloud $\mathcal{P}_k$, previous clouds $\mathcal{P}_{k-1}$, $\mathcal{P}_{k-2}$, threshold $d$, incremental estimates $\mathbf{T}_{k-2, k-1}$, $\mathbf{T}_{k-1, k}$
\State \textbf{Output:} filtered cloud $\mathcal{Q}_{k-1}$ to be inserted to the map
\State \textbf{Initialization:} $\mathcal{Q}_{k-1} \gets \{ \}$
\For {each point $\mathbf{p}_{k-1} \in \mathcal{P}_{k-1}$}
\State transform $\mathbf{p}_{k-1}$ to frame $\{ k \}$ by $\mathbf{T}_{k-1, k}$ to obtain $\mathbf{p}_{k}$ and find its nearest neighbor in $\mathcal{P}_{k}$ as $\mathbf{p}_{k}^{*}$
\State transform $\mathbf{p}_{k-1}$ to frame $\{ k-2 \}$ by $\mathbf{T}_{k-2, k-1}$ to obtain $\mathbf{p}_{k-2}$ and find its nearest neighbor in $\mathcal{P}_{k-2}$ as $\mathbf{p}_{k-2}^{*}$
\If {$\mathbf{p}_{k} - \mathbf{p}_{k}^{*} < d$ \textbf{or} $\mathbf{p}_{k-2} - \mathbf{p}_{k-2}^{*} < d$}
\State $\mathcal{Q}_{k-1} \gets \mathcal{Q}_{k-1} \cup \mathbf{p}_{k-1}$
\EndIf
\EndFor
\end{algorithmic}
\end{algorithm}

\subsubsection{Mapping Consistency Filter}
The Mapping Consistency Filter determines what to update to the map. 
The goal is to select incrementally consistent
points and filter out unreliable points related to translation-induced distortion and sparse dynamic objects (e.g., a nearby safety operator).
As introduced in Algorithm~\ref{algorithm2}, the proposed method takes as input the three recent frames associated with their incremental odometry estimates and produces a filtered cloud of the intermediate frame. 
The key idea is to check if points in the intermediate frame also appear consistently (within a reasonably small distance threshold, e.g., 0.05\;m) in the previous or the next frame. 
Points that do not meet this criterion will be removed.

Figure~\ref{fig_consistency} provides an example of the aforementioned operation principle. As shown, points that are consistently present across consecutive frames (yellow segments overlapping with blue and red segments) are retained for map updates, whereas points that appear only in the intermediate frame (yellow segment distant from the red segment) are removed.
The computational time overhead of this filter is minimal, as it is applied to only the keyframes to be inserted into the map.

\begin{figure}[!t]
\centering
\includegraphics[width=\linewidth]{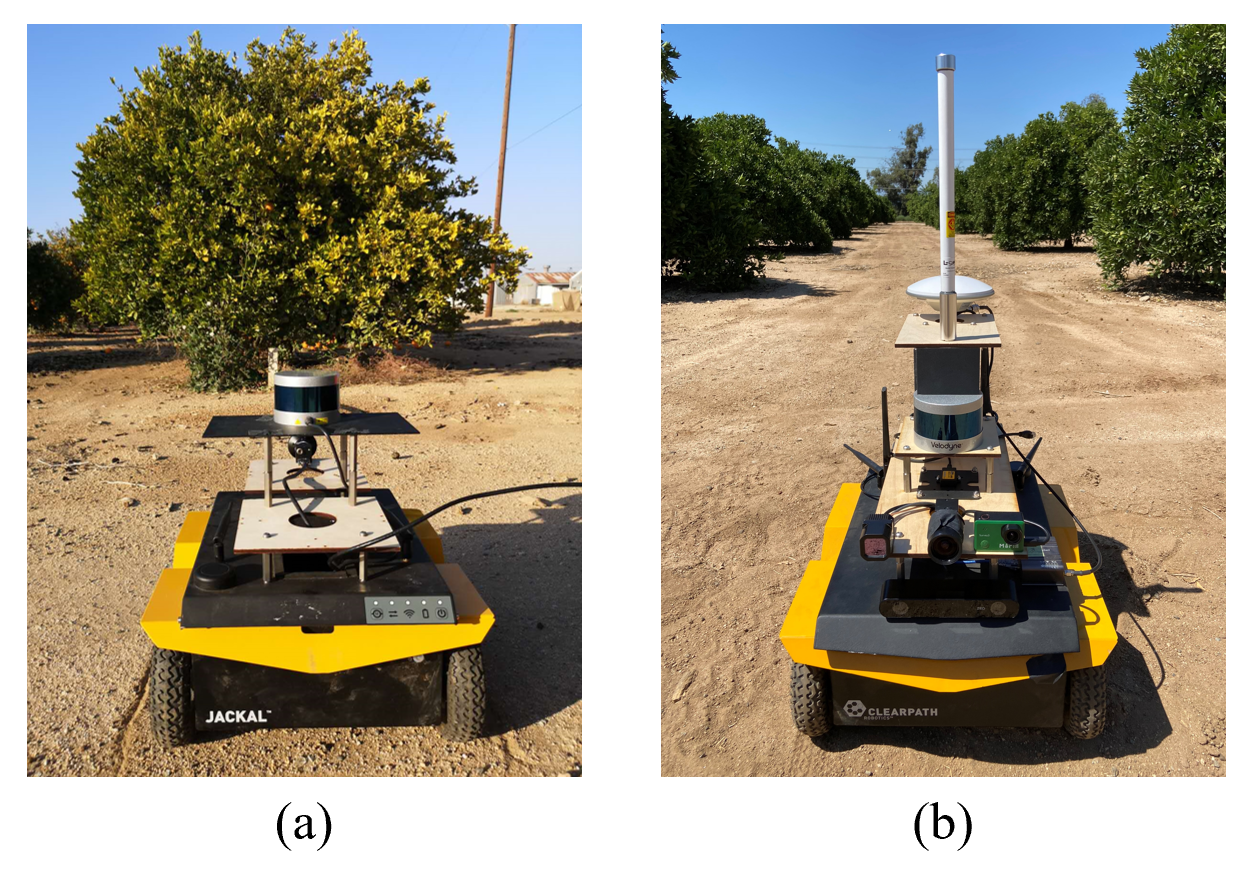}
\caption{The Clearpath Robotics Jackal mobile robot used in this work is shown at its departure (initial) position in the field. (a) In the early-stage design of our sensor payload, only the Velodyne VLP-16 LiDAR sensor was used for data collection (sequences A1-A6 and B1-B5). (b) The latest sensor payload design included LiDAR and GPS-RTK sensors for data collection (sequences C1-C7).}
\label{fig_jackal}
\end{figure}

\begin{table*}[!t]
\caption{Dataset Characteristics}
\label{table_dataset}
\begin{center}
\resizebox{1.9\columnwidth}{!}{
\begin{tabular}{cccccccccc}
\toprule
Seq. & \makecell{Planting \\ Env. } & Terrain & Path Pattern & \makecell{Distance \\ (m)} & \makecell{Duration \\ (min:sec)} & \makecell{Controller}  & \makecell{Return \\ to origin} & GPS-RTK \\
\midrule
A1 & in-row & rugged & single-round & 127.34 & 6:19 & arrow-key & Yes & No \\
A2 & in-row  & rugged & single-round & 250.07 & 11:44 & arrow-key & Yes & No \\
A3 & mixture  & rugged & single-round & 128.21 & 6:03 & arrow-key & Yes & No \\
A4 & mixture & rugged & single-round & 164.17 & 7:25  & arrow-key & Yes & No \\
A5 & uniform  & flat & single-round & 140.01 & 6:53 & arrow-key & Yes & No \\
A6 & uniform  & flat & single-round & 183.92 & 5:12 & arrow-key & Yes & No \\
\midrule
B1 & mixture  & rugged & cross-trees & 434.21 & 8:11 & joystick & Yes & No \\
B2 & mixture  & rugged & cross-trees & 443.89 & 8:50 & joystick & Yes & No \\
B3 & mixture & bumps/dips & lawn-mower & 1162.34 & 16:33  & joystick & Yes & No \\
B4 & uniform  & flat & cross-trees & 736.64 & 12:49 & joystick & Yes & No \\
B5 & uniform  & flat & lawn-mower & 1758.01 & 26:08 & joystick & Yes & No \\
\midrule
C1 & in-row & rugged & single-round & 132.50 & 3:58 & joystick & Yes & Yes \\
C2 & in-row  & bumps/dips & lawn-mower & 644.33 & 12:49 & joystick & Yes & Yes \\
C3 & in-row & bumps/dips & lawn-mower & 390.79 & 7:44 & joystick & Yes & Yes \\
C4 & mixture & bumps/dips & lawn-mower & 408.27 & 7:39  & joystick & Yes & Yes \\
C5 & mixture & rugged & cross-trees & 145.50 & 4:06  & joystick & Yes & Yes \\
C6 & mixture  & rugged & cross-trees & 173.75 & 5:12 & joystick & Yes & Yes \\
C7 & mixture & rugged & cross-trees & 192.87 & 6:38 & joystick & Yes & Yes \\
\bottomrule
\end{tabular}
}
\end{center}
\end{table*}

\begin{figure}[!t]
\centering
\includegraphics[width=\linewidth]{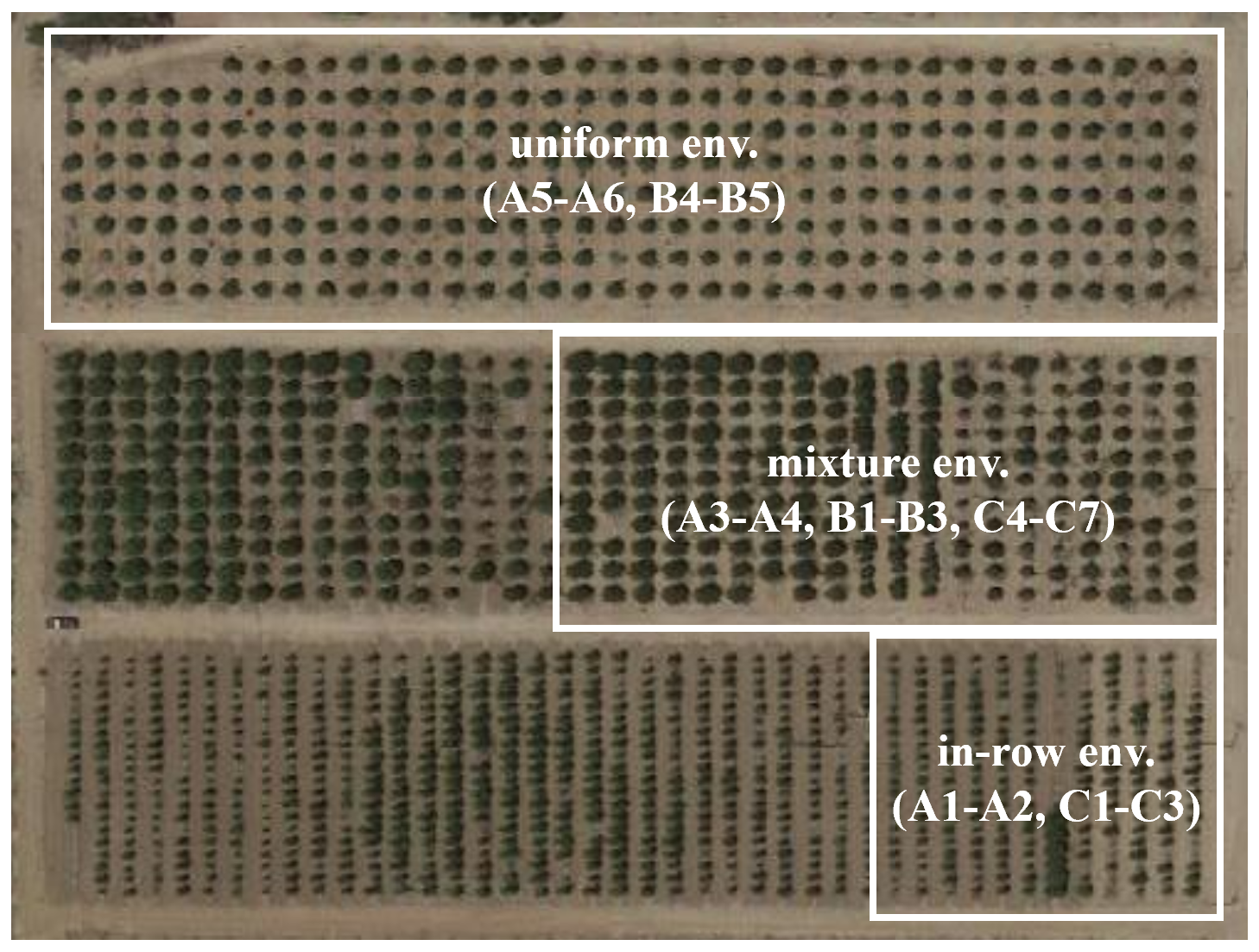}
\caption{Composite satellite image of the agricultural fields where field experiments were conducted (courtesy of Google Maps). Areas corresponding to each sequence are highlighted accordingly.}
\label{fig_satellite}
\end{figure}

\subsection{Experimental Setup}\label{sec_experiment}
We collected experimental data in real-world agricultural fields at the University of California, Riverside (UCR) Agricultural Experimental Station (AES; Figure~\ref{fig_field}) using a Clearpath Jackal mobile robot (Figure~\ref{fig_jackal}).
During data collections, the robot was remotely piloted and followed by a human operator to travel through various fields. 
Table~\ref{table_dataset} summarizes the characteristics of each dataset sequence, and Figure~\ref{fig_satellite} shows satellite imagery of the fields where each sequence was collected.

To better evaluate our algorithm's performance, three distinctive experimental phases were performed. These are shown as sequences (A, B, and C) in Table~\ref{table_dataset}. 
These phases were scheduled sequentially several months apart and with distinct objectives. 
Specifically, the first phase aimed to evaluate the adaptability of the algorithm in diverse environments. 
The data sequences collected in this phase (sequences A1-A6 in Table~\ref{table_dataset}) were short trajectories (e.g., 100-250\;m) that took place in three different agricultural settings in terms of planting environment type (in-row, mixture, uniform) and terrain type (rugged, flat). 
The aim of the second phase (sequences B1-B5 in Table~\ref{table_dataset}) was to evaluate algorithm performance over a range of more challenging scenarios, namely long-distance travel, navigation under the canopy, crossing between trees, and traversing over furrows. 
Compared to the previous phase, we considered different combinations in terms of planting environment type, terrain type, and employed path pattern (Table~\ref{table_dataset}). 
In all experiments considered under the first two phases, the robot was configured as shown in Figure~\ref{fig_jackal}(a), and it was controlled to start and finish at identical locations. 
The third and final phase (sequences C1-C7 in Table~\ref{table_dataset}) aimed to assess the estimation error throughout an entire trajectory. 
For this reason, the robot was equipped with a GPS receiver with Real-time Kinematic Positioning (RTK) to offer ground truth measurements (Figure~\ref{fig_jackal}(b)). 
A GPS base station was placed nearby to communicate with the robot its current GPS location and facilitate the calculation of a global robot pose with about 2\;cm localization accuracy. 
Sequences were collected to maintain the diversity of the environment and path pattern similar to previous phases.
In total, the dataset collected herein spans a total operation time of 2.5\;hrs, covers a total distance of 7.5\;km, and features various combinations of planting environments and robot motion profiles. 

We also compared our proposed adaptive LiDAR-only odometry and mapping framework against five state-of-the-art related approaches: LeGO-LOAM~\cite{shan2018lego}, CT-ICP~\cite{dellenbach2022cticp}, KISS-ICP~\cite{vizzo2023kiss-icp}, HDL-Graph-SLAM~\cite{koide2019hdl} and LOCUS~\cite{palieri2020locus, reinke2022locus}. 
LeGO-LOAM is a feature-based method; CT-ICP and KISS-ICP are based on point-to-plane and point-to-point ICP, respectively; HDL-Graph-SLAM and LOCUS are based on Generalized ICP (i.e. plane-to-plane ICP). 
All evaluations were conducted on the same laptop machine with Ubuntu 20.04 operating system, an i7-12700H CPU and 32\;GB RAM.

\section{Results and Discussion}

\subsection{Assessment of Main Experimental Findings}
\begin{figure*}[!t]
\centering
\includegraphics[width=\linewidth]{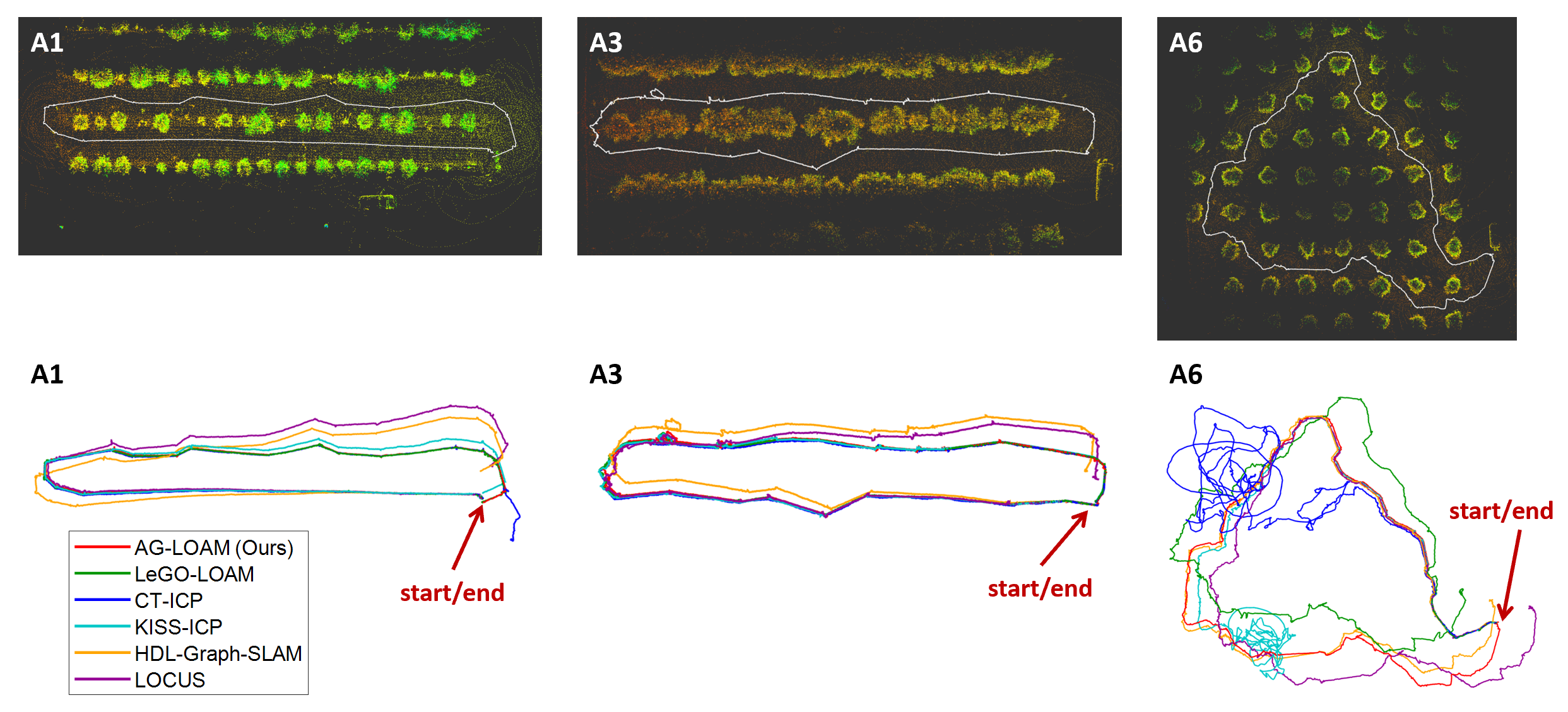}
\caption{Estimated trajectory and simultaneous mapping result of the method developed herein (top panel), and comparison of trajectories (bottom panel) against other methods on sequences A1, A3, A6 in the dataset. (Figure best viewed in color.)
}
\label{fig_map_traj_A}
\end{figure*}

The main experimental findings from field testing are presented and discussed next. 
Quantitative results corresponding to experimental phases A, B, and C are reported in Tables~\ref{table_endpoint_A},~\ref{table_endpoint_B} and~\ref{table_ate}, respectively. 
Corresponding qualitative results are presented in Figure~\ref{fig_map_traj_A} (phase A), Figures~\ref{fig_map_traj_B13} and~\ref{fig_map_traj_B5} (phase B), and Figure~\ref{fig_map_traj_C} (phase C). 

\begin{table*}[!t]
\caption{Quantitative evaluation of data sequences in experiment phase A.
The distance (in meters) from the origin to the endpoint of the trajectory is reported.  
Table entries marked in bold font indicate successful runs that returned to the origin area and produced clear and complete maps.}
\label{table_endpoint_A}
\begin{center}
\resizebox{1.3\columnwidth}{!}{
\begin{tabular}{c|cccccc}
\toprule
Sequence & A1 & A2 & A3 & A4 & A5 & A6 \\
\midrule
LeGO-LOAM  & \textbf{0.51} & 1.06 & \textbf{0.40}  & \textbf{0.53} &  6.23 & 7.51  \\
CT-ICP & 6.24  & 22.74 & \textbf{0.40} & \textbf{0.56} & 33.88 & 57.75 \\
KISS-ICP  & 0.71 & 4.46 & \textbf{0.52} & 2.47 & 29.47 & 37.33 \\
HDL-Graph-SLAM  &  3.62 & 3.11 & 3.98 & 2.43 & \textbf{1.02} & 4.52 \\
LOCUS &  4.06 &  2.18 & 2.97 & 6.15 & 36.00 & 6.06 \\
AG-LOAM (Ours)  & \textbf{0.46} & \textbf{0.67} & \textbf{0.36} & \textbf{0.53} & \textbf{0.71} & \textbf{0.34} \\
\bottomrule
\end{tabular}
}
\end{center}
\end{table*}

\subsubsection{Experimental Phase A: Simple Trajectories in Various Environments}

The evaluation metric we employed is the distance (in meters) from the origin to the endpoint of the trajectory, as in~\cite{shan2018lego}. 
We manually inspected the mapping results and estimated trajectories and marked successful runs that returned to the origin in bold font (Table~\ref{table_endpoint_A}). 
Results demonstrate that our method performs \emph{consistently} well across different planting environments.

Sample qualitative results in in-row, uniform, and mixture planting environments (Figure~\ref{fig_map_traj_A}) demonstrate that our method can provide estimated trajectories that return to the origin successfully and produce accurate maps without blur. 
Other methods demonstrate distortion and accumulated drift at varying levels. 
Notably, CT-ICP is generally well-performing and its estimated trajectories are aligned with our method's ones at most times. 
However, when its frame-to-frame elastic registration fails at any point, the rest of the trajectory can deviate widely and hardly recover (sequence A6). 
This is due to its inherent continuous-time optimization design. 
Similarly, KISS-ICP also suffers from registration failures and this can lead to corrupted estimation in place (sequence A6). 

Experiments in this phase provided two observations on planting environments. 
First, in-row planting environments are generally deemed challenging if trees are very similar to each other and are planted densely since this can block a large field of view and create a corridor-like environment. 
However, if trees are of diverse appearance (e.g., different species, different growth levels, or distinctive pruning patterns) and/or are planted sparsely, unique features from the surrounding environment can facilitate localization (sequences A1-A4). 
Second, uniform planting environments are self-similar and are challenging when attempting sharp turns (sequences A5-A6) because cloud points describing one tree can be easily mismatched with another tree after a 90-degree rotation.

\subsubsection{Experiment Phase B: Diverse Challenging Trajectories}

\begin{figure*}[!t]
\centering
\includegraphics[width=0.78\linewidth]{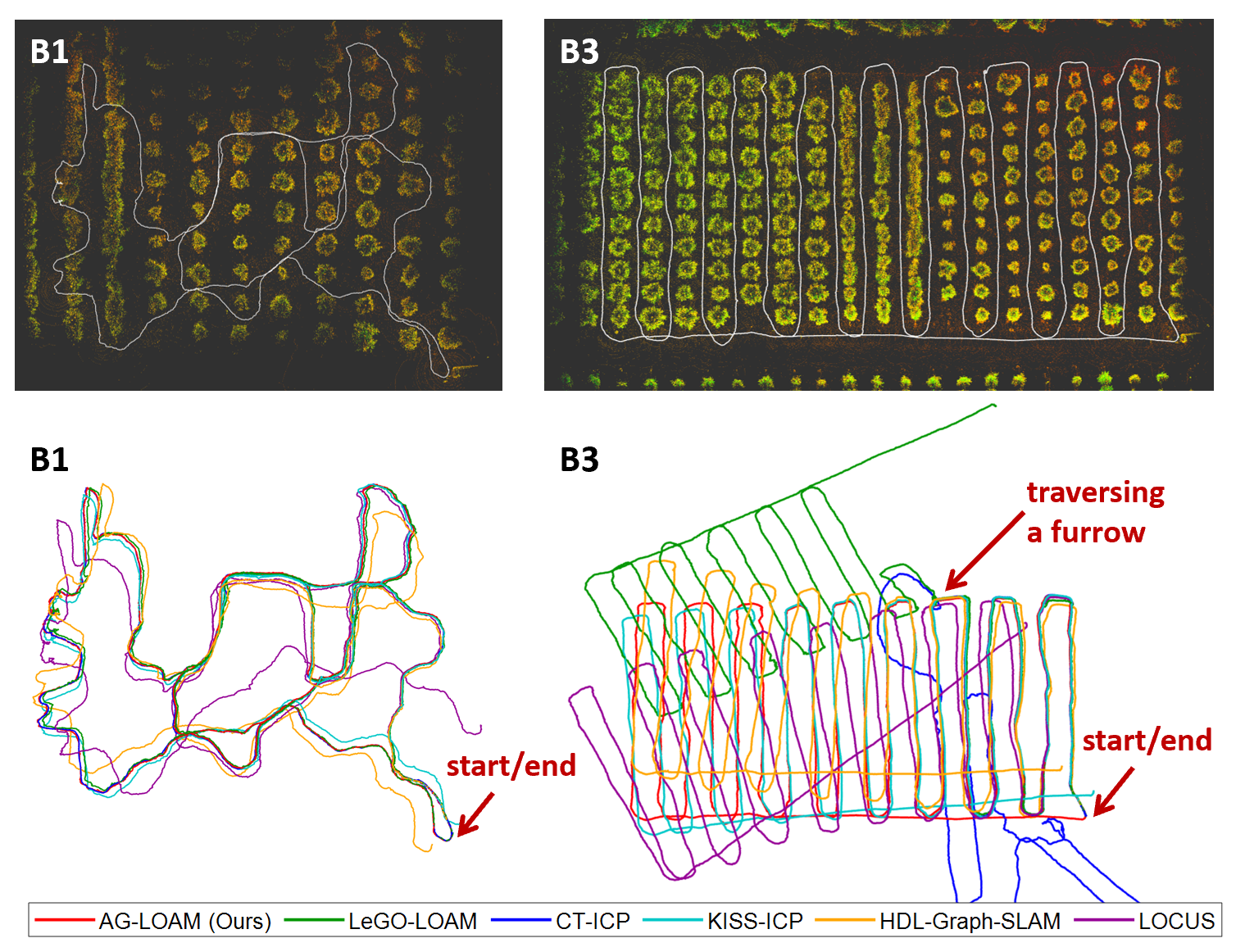}
\caption{Estimated trajectory and simultaneous mapping result of the method developed herein (top panel), and comparison of trajectories (bottom panel) against other methods on sequences B1, B3 in the dataset. (Figure best viewed in color.)
}
\label{fig_map_traj_B13}
\end{figure*}

In sequences B1-B2, we investigated challenging \emph{robot paths} that consist of massive turns, tree-crossing, and navigation under canopies. 
An example of such a robot path is illustrated in the left column of Figure~\ref{fig_map_traj_B13}. 
Most algorithms (including ours) performed well in this case (columns B1 and B2 in Table~\ref{table_endpoint_B}). 
This may be credited to the fact that rich features and unique landmarks are still available under such paths for algorithms to utilize, despite the quite challenging path followed.

\begin{table*}[!h]
\caption{Quantitative evaluation of data sequences in experiment phase B.
The distance (in meters) from the origin to the endpoint of the trajectory is reported. 
Table entries marked in bold font indicate successful runs that returned to the origin area and produced clear and complete maps.}
\label{table_endpoint_B}
\begin{center}
\resizebox{1.2\columnwidth}{!}{
\begin{tabular}{c|ccccc}
\toprule
Sequence & B1 & B2 & B3 & B4 & B5 \\
\midrule
LeGO-LOAM  & \textbf{0.17} & \textbf{0.07} & 111.01  & \textbf{0.14} &  0.70 \\
CT-ICP & \textbf{0.06}  & \textbf{0.05} & 56.79 & 85.00 & 44.34 \\
KISS-ICP  & 2.80 & 1.38 & 6.35 & 23.86 & 43.49 \\
HDL-Graph-SLAM  &  6.95 & 9.27 & 13.79 & 13.07 & 35.48 \\
LOCUS &  15.89 &  22.31 & 54.54 & 59.11 & 64.47 \\
AG-LOAM (Ours)  & \textbf{0.10}  & \textbf{0.03} & \textbf{0.09} & \textbf{0.14} & \textbf{0.14}  \\
\bottomrule
\end{tabular}
}
\end{center}
\end{table*}

\begin{figure*}[!t]
\centering
\includegraphics[width=0.75\linewidth]{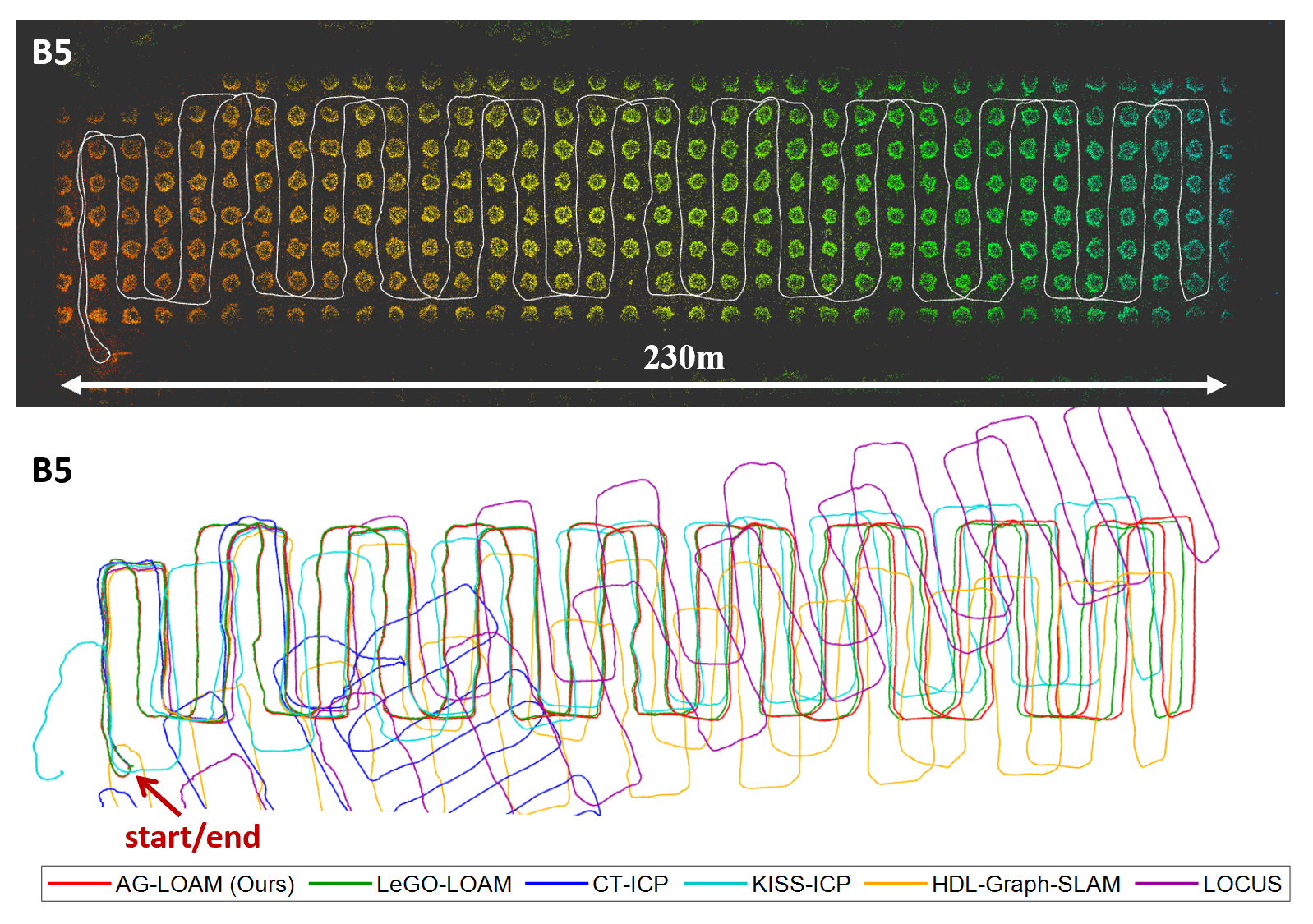}
\vspace{-6pt}
\caption{Estimated trajectory and simultaneous mapping result of the method developed herein (top panel), and comparison of trajectories (bottom panel) against other methods on sequence B5 in the dataset. (Figure best viewed in color.)
}
\label{fig_map_traj_B5}
\vspace{-6pt}
\end{figure*}

Next, in sequence B3, we moved on to consider variable types of \emph{terrain features}, such as bumps and dips, and especially traversing over furrows on the side of the field. 
(An example of a furrow is illustrated in Figure~\ref{fig_field}(d).) 
Results (column B3 in Table~\ref{table_endpoint_B}) demonstrate that few algorithms can handle continuous high-rate pitch motion (ups and downs) in the forward direction. 
Some methods can tackle a few traverses over the furrow but soon deviate widely. 
In contrast, our developed method can robustly provide accurate trajectory and mapping results, owing to the adaptive mapping mechanism.

In sequences B4-B5, we focused on \emph{long-distance travel} to evaluate how cumulative drift can evolve in various methods. 
Qualitative results in Figure~\ref{fig_map_traj_B5} readily demonstrate the superior performance of our method. 
Quantitative results (columns B4 and B5 in Table~\ref{table_endpoint_B}) also indicate that LeGO-LOAM almost succeeds in both cases, potentially owing to a smoother motion profile (we used a joystick controller in seq. B4-B5 as opposed to arrow buttons used in seq. A5-A6). 
Furthermore, it should be noted that the results for LeGO-LOAM were obtained with map visualization turned off; in the opposite case (visualization switched on) there were delays in computation which resulted in increasing the trajectory estimation error.

\subsubsection{Experimental Phase C: Benchmarking Absolute Errors in Robot Translation}

\begin{figure*}[!t]
\centering
\includegraphics[width=\linewidth]{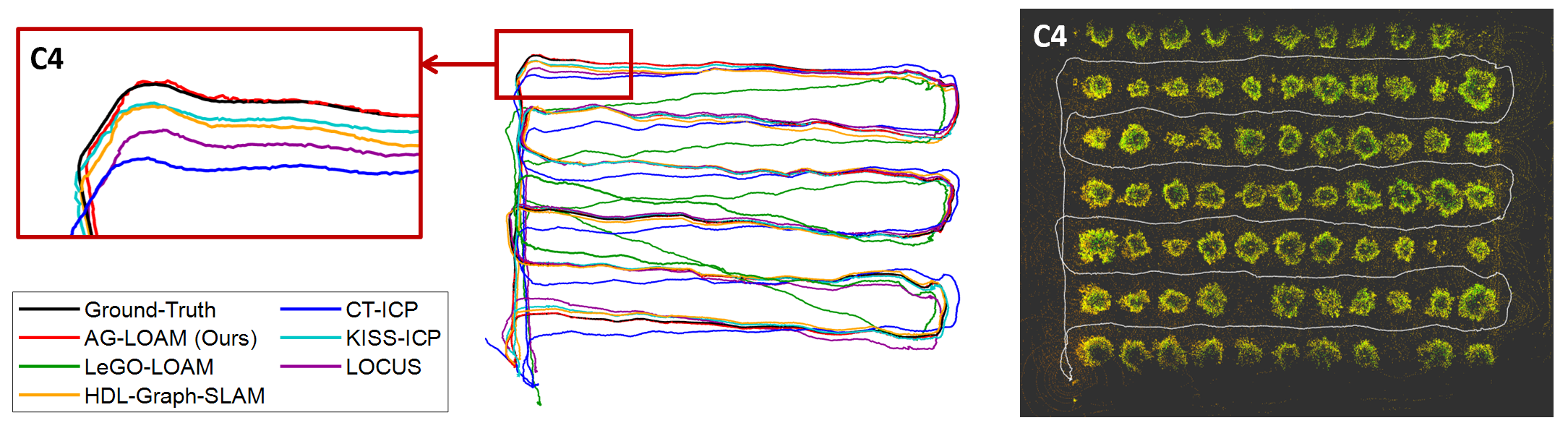}
\caption{Estimated trajectory and simultaneous mapping result of the method developed herein (right panel), and comparison of trajectories (left panel) against other methods on sequence C4 in the dataset. (Figure best viewed in color.)
}
\label{fig_map_traj_C}
\end{figure*}

The evaluation metric employed in this set of experiments was the absolute translational error (as in~\cite{xu2022fastlio2}). 
This metric computes the root mean square error (RMSE) between the estimated trajectory and the ground-truth trajectory for every frame of point cloud available in the data sequence. 
Our method performs \emph{consistently} well across all cases (Table~\ref{table_ate}), with cases C1 and C6 being very close to LeGO-LOAM that yielded the best performance in these two specific sequences. 
This finding is consistent with all other obtained results described earlier and attests to our method's consistent and accurate performance. 
Note that the results presented herein were obtained using the same set of parameters that can work well in general without fine-tuning. 
If needed, parameters can be tuned for a specific data sequence to further improve performance. 
Figure~\ref{fig_map_traj_C} offers an example of the estimated trajectories on sequence C4, with a close-up view of the comparison to the ground-truth trajectory.  
Both quantitative and qualitative results demonstrate that our method achieves minimal estimation errors over the entire trajectory in comparison to other methods.

\begin{table*}[!t]
\caption{Quantitative evaluation of data sequences in experiment phase C.
The absolute translational error (RMSE, in meters) computed over the entire trajectory is reported. 
Table entries marked in bold font indicate the best results.}
\label{table_ate}
\vspace{-5pt}
\begin{center}
\resizebox{1.5\columnwidth}{!}{
\begin{tabular}{c|ccccccc}
\toprule
Sequence & C1 & C2 & C3 & C4 & C5 & C6 & C7 \\
\midrule
LeGO-LOAM & \textbf{0.065} & 0.155 & 0.092 & 6.475 & 0.126 & \textbf{0.070} & 0.123 \\
CT-ICP  &  0.091 &  0.118 &  0.114 &  2.885 &  1.372 &  0.088 &  0.138 \\
KISS-ICP   &  0.135 &  1.152 &  0.405 &  0.650 &  0.284 &  0.194 &  0.121 \\
HDL-Graph-SLAM &  1.153 &  1.687 &  1.259 &  1.020 &  0.666 &  0.795 &  0.705 \\
LOCUS  &  0.694 &  1.932 &  2.730 &  0.957 &  0.156 &  0.146 &  0.233 \\
AG-LOAM (Ours) &  0.070 &  \textbf{0.116} &  \textbf{0.091} &  \textbf{0.092} &  \textbf{0.092} &  0.103 &  \textbf{0.075} \\
\bottomrule
\end{tabular}
}
\end{center}
\end{table*}

\subsubsection{Summarized Discussion from Findings Across all Cases}
Further, we briefly discuss the possible reasons why other methods may be underperforming in different cases. 
KISS-ICP maintains only a local map and therefore suffers from cumulative errors that are inevitable in long-distance travel. 
As GICP-based dense methods, both HDL-Graph-SLAM and LOCUS can achieve accurate estimation in translational motion, but can suffer from high-rate rotations as IMU is unavailable for correcting distortion and their current map updating approaches are incapable of removing such distorted points, thus resulting in drift in estimated trajectories. 
LeGO-LOAM alleviates distortion by segmenting point clouds and removing ground points before feature matching. 
This approach works well in feature-rich environments (seq. A1-A4, B1-B2) and smooth motion (seq. B4-B5), but suffers from self-similar environments (seq. A5-A6) and when robot motion is more abrupt (seq. B3, C4). 
In contrast, our proposed method addresses distortion and unreliable points using an adaptive mapping mechanism and achieves accurate and consistent performance across all field conditions.

\subsection{Evaluation of Computational Efficiency}\label{sec_computation}
\begin{figure}[!h]
\begin{center}
\includegraphics[width=0.85\linewidth]{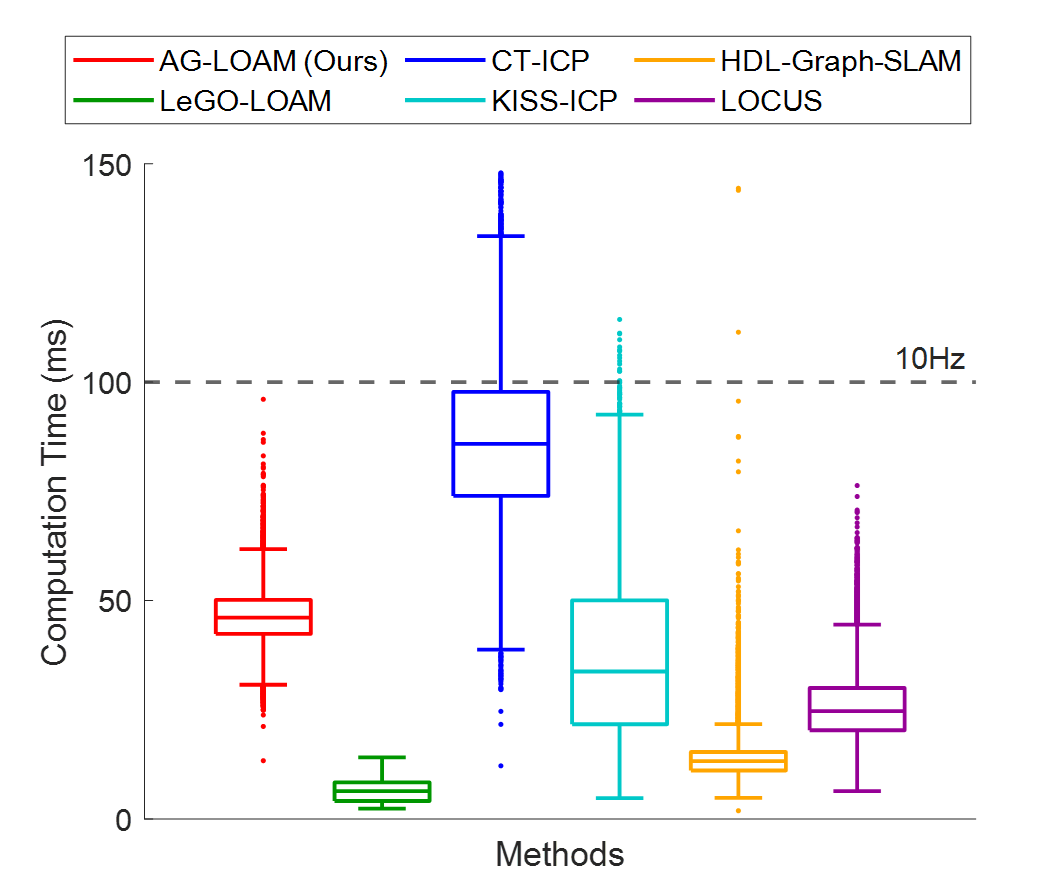}
\caption{Boxplot of the computation time (ms) for each method evaluated on the B5 sequence in the dataset. (Figure best viewed in color.)}
\label{fig_time}
\end{center}
\end{figure}

\begin{figure}[!h]
\begin{center}
\includegraphics[width=0.85\linewidth]{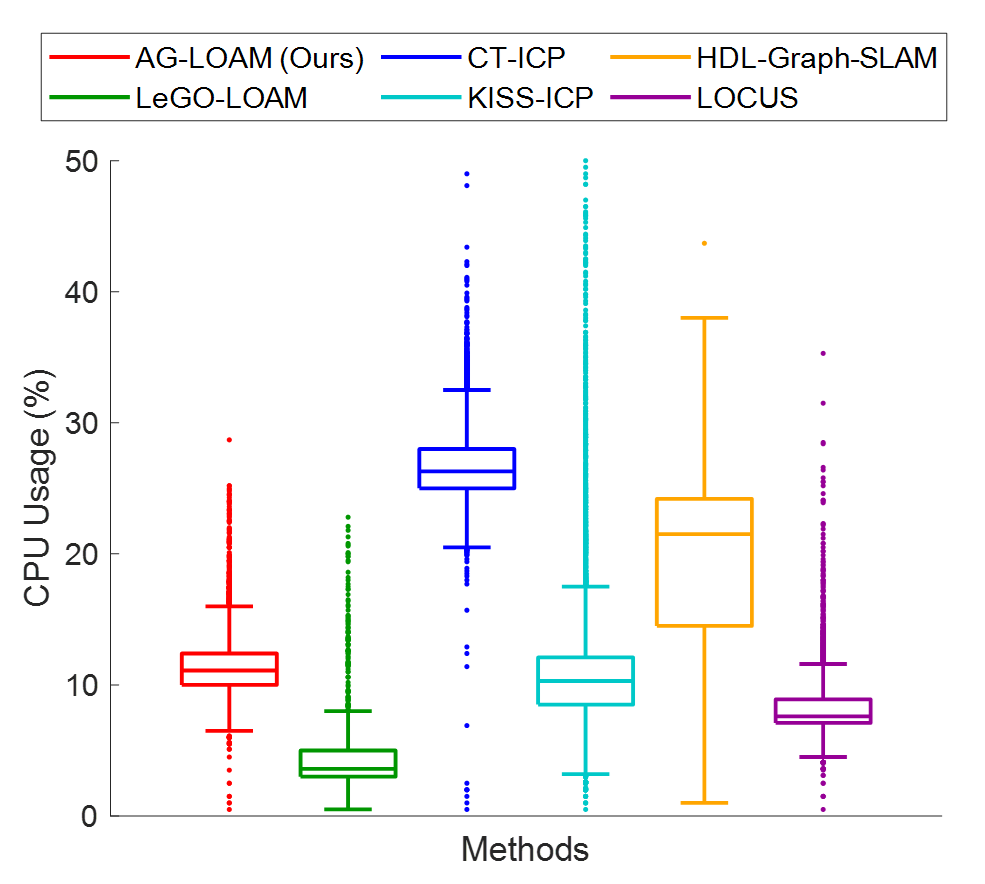}
\caption{Boxplot of the CPU usage (\%) for each method evaluated on the B5 sequence in the dataset. (Figure best viewed in color.)}
\label{fig_cpu}
\end{center}
\end{figure}

The computational efficiency of each method was evaluated in terms of two aspects: 1) the computation time needed for the algorithm to provide an odometry estimate, timed from the moment a new point cloud frame was received; and 2) the percentage of CPU usage over the trajectory when the algorithm is running. 
We ran all methods on the sequence that features long-distance travel (B5 in Table~\ref{table_dataset}). 
Time consumption and CPU usage were recorded for each point cloud frame, constituting about 15,000 data samples. 
Obtained computation time and CPU usage results are reported in Figure~\ref{fig_time} and Figure~\ref{fig_cpu}, respectively. 
Results indicate that most methods can achieve real-time execution at 10Hz, and the CPU usage of our proposed method is competitive compared to others.

\subsection{Study of the Effect of Key Settings and Parameters in Algorithm Performance}\label{sec_ablation}
Lastly yet importantly, we present results from assessing how different key settings and parameters of our method affect its performance. 
First, to assess the value of different components in the proposed algorithm framework, we conducted an ablation study regarding the two mapping modules: the Motion Stability Filter and the Mapping Consistency Filter. 
Second, to evaluate the effect of key parameters employed in the mapping algorithm, we conducted another ablation study centered on three parameters: the minimum travel distance $t$ (in meters) required for map updates, the maximum rotation $\theta$ (in degrees) allowed for map updates, and the maximum displacement $d$ (in meters) allowed for consistent points. 
The first two parameters decide when to update the map while the last parameter determines what to update in the map.

\begin{table}[t]
\caption{Ablation study on the proposed Motion Stability Filter and Mapping Consistency Filter. 
The distance (in meters) from the origin to the endpoint of the trajectory is reported.
Values marked in bold font indicate successful runs that returned to the origin area and produced clear and complete maps.
}
\label{table_ablation}
\begin{center}
\resizebox{\columnwidth}{!}{
\begin{tabular}{c|ccccc} 
\toprule
Sequence & A5 & A6 & B1 & B3 & B5 \\
\midrule
AG-LOAM (w/o both filters) & 1.45 & 5.86 & 0.87 & 21.78  & 1.97 \\
AG-LOAM (w/o consistency filter) & \textbf{0.69} & 5.51 & \textbf{0.09} & 16.54 & \textbf{0.13} \\ 
AG-LOAM (w/o stability filter)  & \textbf{0.71} & \textbf{0.34} & \textbf{0.10} & 2.55 & \textbf{0.14} \\ 
AG-LOAM  & \textbf{0.71} & \textbf{0.34} & \textbf{0.10} & \textbf{0.09} & \textbf{0.14} \\ 
\bottomrule
\end{tabular}
}
\end{center}
\end{table}

\begin{figure}[!t]
\centering
\includegraphics[width=\linewidth]{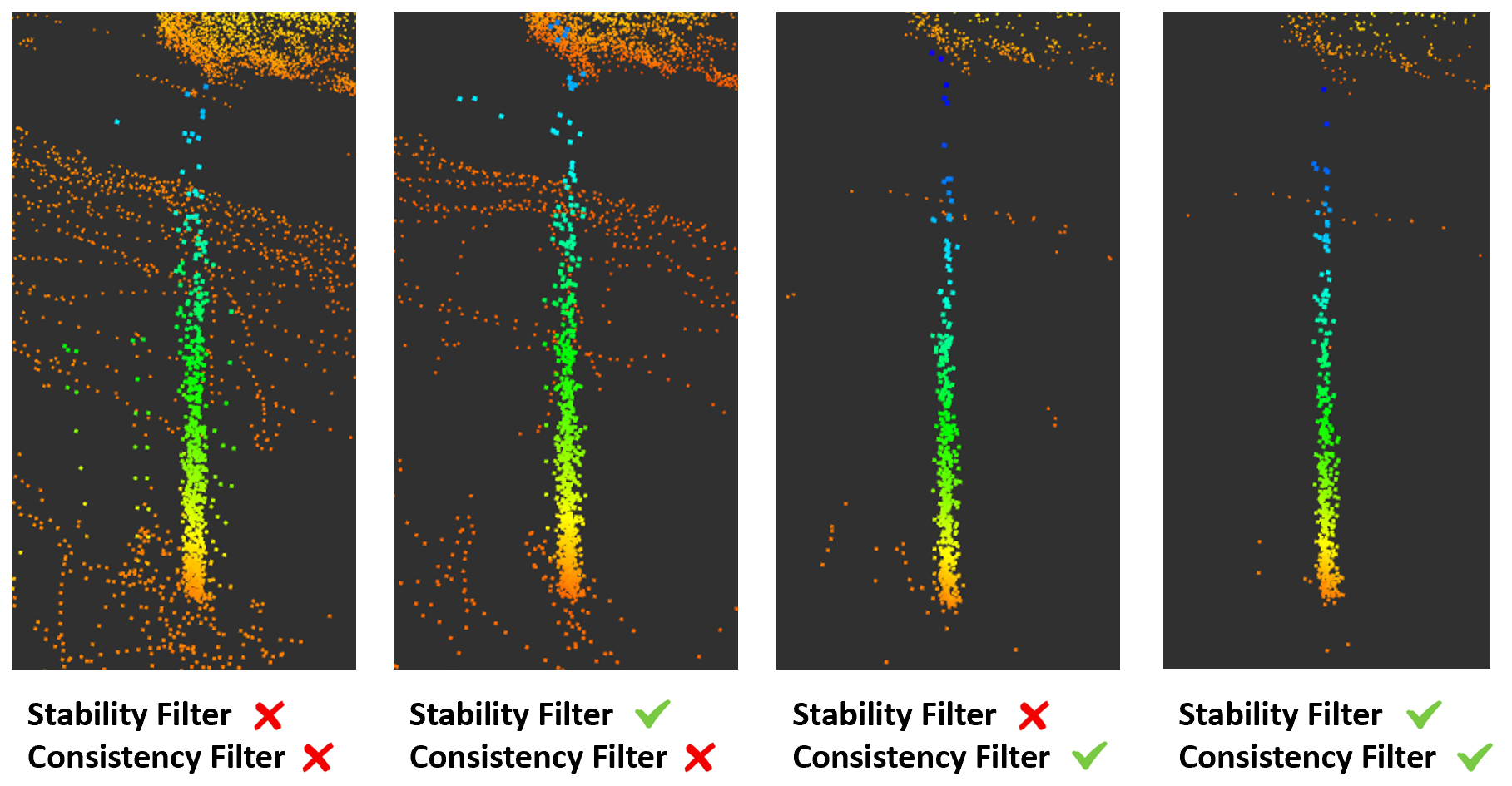}
\caption{Ablation study on the proposed Motion Stability Filter and Mapping Consistency Filter. This qualitative evaluation presents a close-up view of a utility pole located on the side of the field in the in-row environment (seq. A1-A2).}
\label{fig_ablation_seq2}
\end{figure}

\begin{figure}[!t]
\centering
\includegraphics[width=\linewidth]{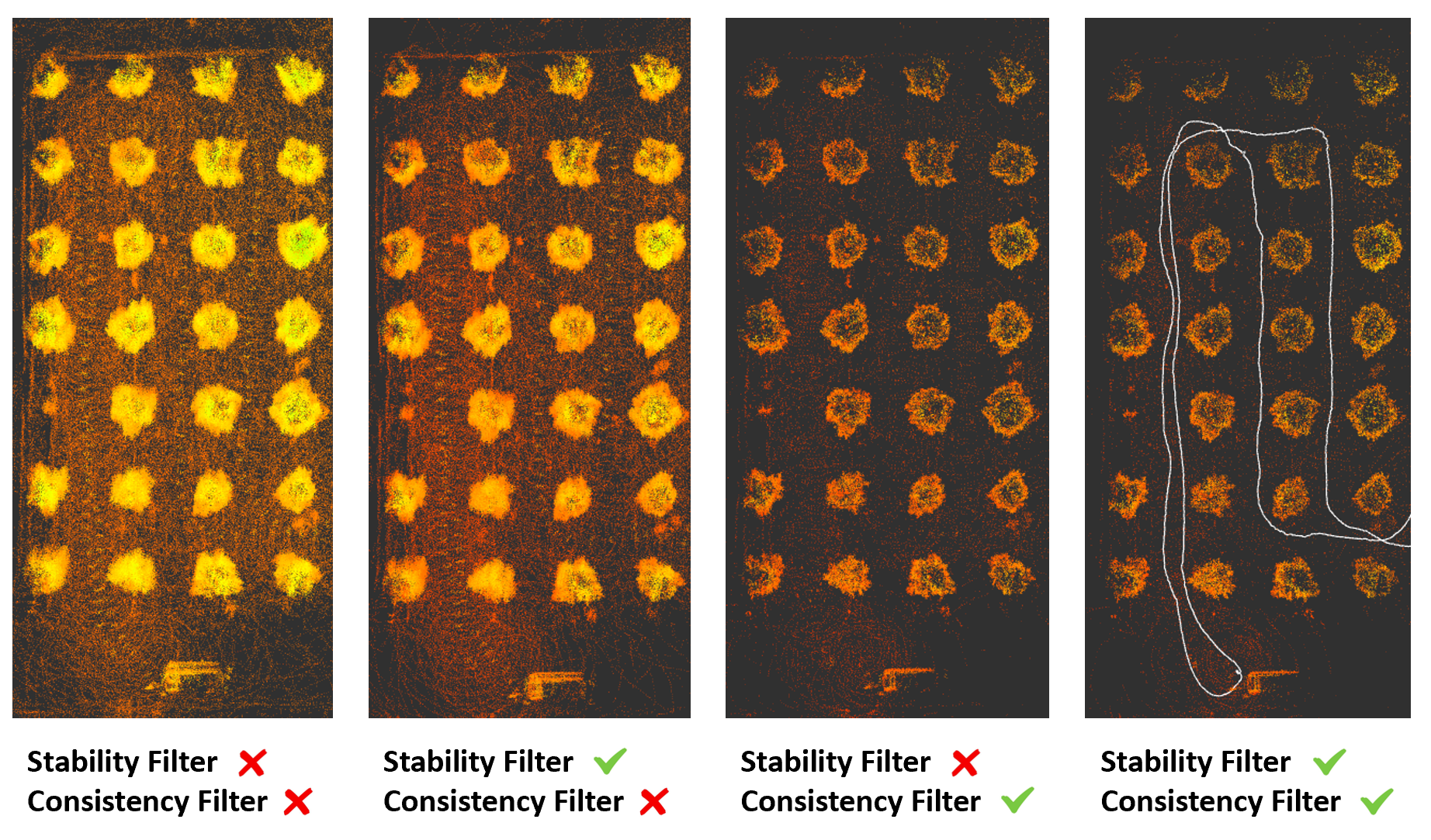}
\caption{Ablation study on the proposed Motion Stability Filter and Mapping Consistency Filter. This qualitative evaluation presents a close-up view of the mapping result in the uniform environment (seq. B5). The estimated trajectory is superimposed on the last panel to aid in locating the human operator's shadows in previous panels.}
\label{fig_ablation_seq11}
\end{figure}

We begin with the quantitative results reported in Table~\ref{table_ablation}, which compared the performance of our proposed method when 1) both filters are disabled, 2) only the Motion Stability Filter is enabled, 3) only the Mapping Consistency Filter is enabled, and 4) both filters are enabled (i.e. the setting proposed in our method). 
Results confirm the effectiveness and utility of the two filters. 
Results also indicate that 1) applying solely the Motion Stability Filter is not capable of addressing sparse dynamic objects (seq. A6, B3), and 2) the Mapping Consistency Filter is able to reduce rotation-induced distortion (as an alternative to Motion Stability Filter) while removing inconsistent dynamic objects (seq. A5, B1, B5). 
In addition, we observed mild blur in mapping results of other sequences not reported in Table~\ref{table_ablation} when both filters are disabled, though their endpoint distances to the origin do not change much.

We then provide qualitative results of our method with and without the two filters enabled in Figure~\ref{fig_ablation_seq2} and Figure~\ref{fig_ablation_seq11}, which help explain the rationale behind our approach to develop the Motion Stability Filter and Mapping Consistency Filter, respectively.
In Figure~\ref{fig_ablation_seq2}, we can observe that the Motion Stability Filter effectively prevents map updates in high-rate rotation motion when distorted points are present, and the Mapping Consistency Filter removes inconsistent points in map updates. 
In Figure~\ref{fig_ablation_seq11}, we can observe that the Motion Stability Filter alleviates rotation-induced distortion, but is incapable of preventing translation-induced distortion or mismatches between previously registered shadows of a human operator in the map and new shadows of the same operator during subsequent visits to the same route. 
In contrast, the Mapping Consistency Filter can effectively eliminate human operator shadows in map updates, thereby preventing incorrect registrations during future visits.

Finally, we discuss the results of the ablation study centered on three key parameters employed in the mapping algorithm. 
These are reported in Tables~\ref{table_param_trans},~\ref{table_param_rot} and~\ref{table_param_consistency}. 
In each study, two parameters were set to a fixed value, and one parameter was manipulated within a reasonable range. 
We can observe that large errors occurred on the boundary of value ranges in Table~\ref{table_param_trans} and Table~\ref{table_param_consistency} (i.e. $t=0.5$\;m on C2 and $t=2.0$\;m on C2 and C4; $d=0.03$\;m on C2, C4, C5 and $d=0.20$\;m on C2, C4). 
In Table~\ref{table_param_rot}, no significant errors were observed on the boundary cases, but the best results were obtained when $\theta$ was set between $2.0\degree$ and $4.0\degree$. 
In summary, the proposed algorithm is not contingent on a specific parameter set, so long as they are within a reasonable range.

\begin{table*}[!t]
\caption{Parametric study of the minimum travel distance $t$ (in meters) in map updates when the other two parameters are fixed at $\theta=2.0\degree$ and $d=0.05$\;m.
The absolute translational error (RMSE, in meters) computed over the entire trajectory on sequences C1-C7 is reported.}
\label{table_param_trans}
\vspace{-5pt}
\begin{center}
\resizebox{1.4\columnwidth}{!}{
\begin{tabular}{c|ccccccc}
\toprule
Sequence & C1 & C2 & C3 & C4 & C5 & C6 & C7 \\
\midrule
$t = 0.5$\;m &  0.070 &  0.248 &  0.102 &  0.119 &  0.089 &  \textbf{0.077} &  \textbf{0.071} \\
$t = 1.0$\;m &  0.066 &  0.202 &  0.117 &  0.138 &  \textbf{0.075} &  0.088 &  0.090 \\
$t = 1.5$\;m &  \textbf{0.065} &  \textbf{0.134} &  0.108 &  \textbf{0.085} &  0.095 &  0.108 &  0.162 \\
$t = 2.0$\;m &  0.097 &  0.839 &  \textbf{0.099} &  0.254 &  0.077 &  0.097 &  0.156 \\
\bottomrule
\end{tabular}
}
\end{center}
\end{table*}

\begin{table*}[!t]
\caption{Parametric study of the maximum rotation $\theta$ (in degrees) allowed for map updates when the other two parameters are fixed at $t=1.0$\;m and $d=0.05$\;m.
The absolute translational error (RMSE, in meters) computed over the entire trajectory on sequences C1-C7 is reported.}
\label{table_param_rot}
\vspace{-5pt}
\begin{center}
\resizebox{1.4\columnwidth}{!}{
\begin{tabular}{c|ccccccc}
\toprule
Sequence & C1 & C2 & C3 & C4 & C5 & C6 & C7 \\
\midrule
$\theta=1.0\degree$ &  0.088 &  0.138 &  0.116 &  0.099 &  0.093 &  0.110 &  0.118 \\
$\theta=2.0\degree$ &  \textbf{0.066} &  0.202 &  0.117 &  0.138 &  0.075 &  \textbf{0.088} &  0.090 \\
$\theta=3.0\degree$ &  0.070 &  \textbf{0.116} &  \textbf{0.091} &  \textbf{0.092} &  0.092 &  0.103 &  \textbf{0.075} \\
$\theta=4.0\degree$ &  0.074 &  0.124 &  0.097 &  0.156 &  \textbf{0.071} &  0.100 &  0.077 \\
$\theta=5.0\degree$ &  0.073 &  0.128 &  0.094 &  0.108 &  0.072 &  0.093 &  0.090 \\
\bottomrule
\end{tabular}
}
\end{center}
\end{table*}

\begin{table*}[!t]
\caption{Parametric study of the maximum displacement $d$ (in meters) allowed for consistent points when the other two parameters are fixed at $t=1.0$\;m and $\theta=2.0\degree$.
The absolute translational error (RMSE, in meters) computed over the entire trajectory on sequences C1-C7 is reported.}
\label{table_param_consistency}
\vspace{-5pt}
\begin{center}
\resizebox{1.4\columnwidth}{!}{
\begin{tabular}{c|ccccccc}
\toprule
Sequence & C1 & C2 & C3 & C4 & C5 & C6 & C7 \\
\midrule
$d=0.03$\;m &  0.074 &  1.773 &  \textbf{0.098} &  0.357 &  2.103 &  0.110 &  0.222 \\
$d=0.05$\;m &  \textbf{0.066} &  0.202 &  0.117 &  \textbf{0.138} &  \textbf{0.075} &  0.088 &  0.090 \\
$d=0.08$\;m &  0.075 &  \textbf{0.161} &  0.101 &  0.196 &  0.083 &  \textbf{0.076} &  0.084 \\
$d=0.10$\;m &  0.077 &  0.219 &  0.113 &  0.160 &  0.083 &  0.085 &  0.076 \\
$d=0.15$\;m &  0.080 &  0.237 &  0.147 &  0.231 &  0.080 &  0.083 &  \textbf{0.069} \\
$d=0.20$\;m &  0.096 &  0.446 &  0.177 &  0.266 &  0.081 &  \textbf{0.076} &  0.077 \\
\bottomrule
\end{tabular}
}
\end{center}
\end{table*}

\subsection{Study of Algorithm Generalizability on Public Datasets}\label{sec_general}
To better validate the effectiveness of our method across diverse environments and sensor setups, we performed additional evaluation on the TreeScope dataset~\cite{cheng2024treescope}. 
Specifically, we evaluated on the UCM-0523 data sequences collected in a pistachio orchard, where both mobile robots and aerial vehicles were employed for data collection, equipped with Ouster LiDAR OS0-128 and OS1-64, respectively. 
The pseudo ground truth for this dataset was provided as the output of applying the LiDAR-inertial odometry method Faster-LIO~\cite{bai2022fasterlio}. 
Results on these sequences can demonstrate the generalizability of our proposed method on different types of LiDAR and agricultural environments.

\begin{table*}[!t]
\caption{Quantitative evaluation of the UCM-0523 data sequences on the TreeScope dataset~\cite{cheng2024treescope}.
The absolute translational error (RMSE, in meters) computed over the entire trajectory is reported. 
Note that this error is measured relative to the pseudo ground truth trajectory provided by Faster-LIO~\cite{bai2022fasterlio}. 
The smallest values are highlighted in bold, indicating the closest trajectory estimation to that of the LiDAR-inertial method Faster-LIO.}
\label{table_treescope}
\begin{center}
\resizebox{1.9\columnwidth}{!}{
\begin{tabular}{c|c|ccccccccc}
\toprule
Sequence & M-01 & U-01 & U-02 & U-03 & U-04 & U-05 & U-06 & U-07 & U-08 & U-09 \\
\midrule
LeGO-LOAM & \textbf{0.085} & 0.968 & 3.213 & 0.677 & 0.170 & 0.417 & 0.154 & 0.156 & 0.178 & 31.923 \\
CT-ICP  & 0.122 & \textbf{0.081} & 0.078 & 0.551 & \textbf{0.115} & \textbf{0.313} & 0.202 & 0.211 & 0.163 & 0.313 \\
KISS-ICP & 0.127 & 0.100 & 0.097 & 0.629 & 0.135 & 0.606 & 0.127 & 0.138 & \textbf{0.154} & \textbf{0.221} \\
HDL-Graph-SLAM & 0.742 & 0.317 & 0.214 & 1.788 & 0.537 & 2.965 & 0.703 & 0.837 & 1.585 & 1.304 \\
LOCUS  & 0.399 & 0.290 & 0.442 & 7.120 & 1.855 & 7.374 & 2.772 & 4.645 & 2.054 & 4.920 \\
AG-LOAM (Ours) & 0.154 & 0.089 & \textbf{0.077} & \textbf{0.483} & 0.182 & 0.368 & \textbf{0.118} & \textbf{0.125} & 0.158 & 0.228 \\
\bottomrule
\end{tabular}
}
\end{center}
\end{table*}

\begin{figure*}[!t]
\centering
\includegraphics[width=0.85\linewidth]{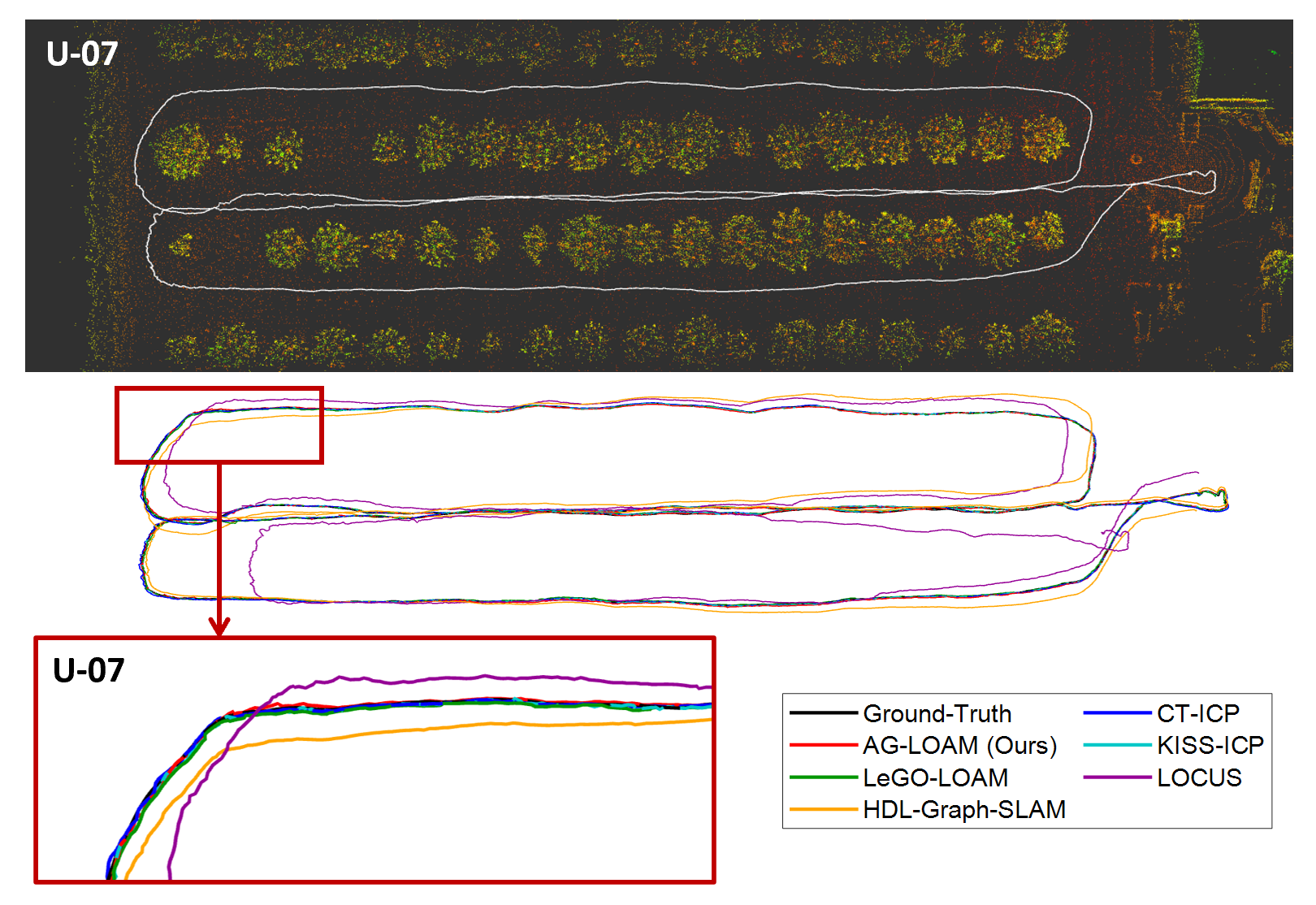}
\caption{Estimated trajectory and mapping result of our method AG-LOAM (top panel), and comparison of acquired trajectories (bottom panel) against other methods on sequence UCM-0523U-07 in the TreeScope dataset. (Figure best viewed in color.)
}
\label{fig_treescope}
\end{figure*}

Table~\ref{table_treescope} contains the quantitative results of our method in comparison to the same five baselines considered earlier.  
Results show that our method consistently produces estimated trajectories that are very close to the results of Faster-LIO~\cite{bai2022fasterlio}, while other methods either fail or deviate significantly from the Faster-LIO results in some cases. 
In cases where our method does not achieve the best performance, our results remain within twice the error of the best-performing method.
Although CT-ICP achieves seemingly good results, it failed to complete the entire trajectory on half of the sequences, even when given twice the amount of time; the current results for CT-ICP are computed from these incomplete estimated trajectories. 
Figure~\ref{fig_treescope} provides a qualitative comparison of all methods, visually confirming the observation that our estimated trajectories are consistently very close to the results of the LiDAR-inertial odometry method Faster-LIO.

\section{Conclusion}
In this work, we introduced an adaptive LiDAR odometry and mapping algorithm for mobile robots operating in complex agricultural environments.
Our proposed framework relies solely on a single 3D LiDAR sensor, making it broadly applicable to a wide range of autonomous agricultural robots and vehicles.
To address the challenge of motion distortion and sparse dynamic objects presenting in unstructured agricultural fields, we proposed an adaptive mapping mechanism that considers motion stability and incremental mapping consistency, which can dynamically adjust when and what to update in the map. 

Experiments were conducted in three phases to cover various planting environments, terrain types and path patterns. 
Results demonstrated that our developed method can provide accurate odometry estimation and mapping results consistently and robustly across diverse settings, whereas other related methods are sensitive to environmental changes and abrupt robot motion. 
Moreover, the computational efficiency of our method is competitive against other methods, making this dense matching method capable of running in real-time on mobile platforms. 
Additional experiments on public datasets validated the generalizability of our method, reaffirming its consistent performance in providing accurate odometry estimates across varying agricultural environments.

One limitation of the current framework is its reliance on the 360-degree field of view of the LiDAR sensor. This can be mitigated by adapting the map update policy to account for rotational changes, and by enhancing the robustness of the scan matching algorithm to handle low-overlap point clouds.
Other potential future directions can include the incorporation of multispectral cameras for multi-modal mapping and crop monitoring (e.g.,~\cite{teng2023multimodal}), and the integration of motion and task planning techniques (e.g.,~\cite{kan2020online, kan2021ral}) for fully autonomous operations in agricultural fields.

\bibliographystyle{IEEEtran}
\bibliography{cas-refs}

\end{document}